\documentclass[11pt]{article}
\usepackage[final]{acl}
\usepackage{times}
\usepackage{latexsym}
\usepackage[T1]{fontenc}
\usepackage[utf8]{inputenc}
\usepackage{microtype}
\usepackage{inconsolata}
\usepackage{amssymb}
\usepackage{amsmath}
\usepackage{amsthm}
\usepackage{graphicx}
\usepackage{subcaption}
\usepackage{xspace}
\usepackage[table]{xcolor} 
\usepackage{tcolorbox}
\usepackage{adjustbox}
\usepackage[ruled]{algorithm2e}
\usepackage{booktabs}
\usepackage{multirow}

\newcommand{\sysnamebase}{AAPO}
\newcommand{\sysname}{\sysnamebase\xspace}
\definecolor{myPurple}{rgb}{0.5, 0, 0.5}
\definecolor{myRed}{rgb}{0.7, 0, 0}
\definecolor{myGreen}{rgb}{0, 0.5, 0}
\definecolor{mygray}{gray}{0.9}
\newtheorem{theorem}{Theorem}

\SetKwFor{ForEach}{for}{do}{end}
\title{\sysname: Enhancing the Reasoning Capabilities of LLMs \\with Advantage Margin}

\author{
 \textbf{Jian Xiong\textsuperscript{1,2}\thanks{This work was done when the first author was an intern in Frontier Research Department, Baidu Inc., under the supervision of Jingbo Zhou.}},
 \textbf{Jingbo Zhou\textsuperscript{2}\thanks{Corresponding authors.}},
 \textbf{Jingyong Ye\textsuperscript{1}},
 \textbf{Qiang Huang\textsuperscript{1}},
 \textbf{Dejing Dou\textsuperscript{1,3}\footnotemark[2]}
\\
 \textsuperscript{1}College of Computer Science and Artificial Intelligence, Fudan University,\\
 \textsuperscript{2}Frontier Research Department, Baidu Inc.,
 \textsuperscript{3}BEDI Cloud
 \\
 {\small \{jxiong24, jyye21, huangq25\}@m.fudan.edu.cn, zhoujingbo@baidu.com, doudejing@fudan.edu.cn}
}

\begin{document}
\maketitle
\begin{abstract}
Reinforcement learning (RL) has emerged as an effective approach for enhancing the reasoning capabilities of large language models (LLMs), especially in scenarios where supervised fine-tuning (SFT) falls short due to limited chain-of-thought (CoT) data.
Among RL-based post-training methods, group relative advantage estimation, as exemplified by Group Relative Policy Optimization (GRPO), has attracted considerable attention for eliminating the dependency on the value model, thereby simplifying training compared to traditional approaches like Proximal Policy Optimization (PPO).
However, existing group relative advantage estimation method still suffers from training inefficiencies, particularly when the estimated advantage approaches zero.
To address this limitation, we propose Advantage-Augmented Policy Optimization (\sysname), a novel RL algorithm that optimizes the cross-entropy (CE) loss using advantages enhanced through a margin-based estimation scheme.
This approach effectively mitigates the inefficiencies associated with group relative advantage estimation.
Experimental results on multiple mathematical reasoning benchmarks and model series demonstrate the superior performance of \sysname.
Code is available at \url{https://github.com/JianxXiong/AAPO}.
\end{abstract}

\section{Introduction}
Reinforcement learning (RL) has emerged as a powerful approach for enhancing the reasoning and decision-making capabilities of large language models (LLMs). While LLMs have demonstrated strong performance in both language understanding and generation tasks \citep{brown2020language, chowdhery2023palm, touvron2023llama1, zhao2023survey}, traditional training strategies such as pre-training and supervised fine-tuning (SFT) \citep{radford2018improving, bommasani2021opportunities, liu2023pre} often fall short in enabling effective chain-of-thought (CoT) \citep{wei2022chain} reasoning for complex decision-making tasks.
To address this limitation, recent research has explored RL-based training paradigms, which have shown considerable empirical success in specialized domains such as mathematical reasoning.
Models including GPT-o1 \citep{openai2024o1}, DeepSeek-R1 \citep{guo2025deepseek}, and QwQ \citep{qwen2024qwq32b} exemplify this promising direction, demonstrating the potential of RL to substantially improve the reasoning capabilities of LLMs.

A recent advancement in LLM post-training with RL is the introduction of novel methods for advantage estimation, a technique for quantifying how favorable a specific action is in a given state.
In this context, the term advantage measures the relative benefit of an action compared to the average in that state, thereby providing an informative learning signal during training.
Traditionally, approaches such as Proximal Policy Optimization (PPO) \citep{schulman2017proximal}, exemplified by InstructGPT \citep{ouyang2022training}, rely on a value model to estimate the advantage.
Although PPO offers stable and reliable performance, maintaining a separate value model leads to substantial consumption of GPU resources.

In contrast to conventional approaches, the group relative advantage estimation method was originally proposed in Group Relative Policy Optimization (GRPO) \citep{shao2024deepseekmath} and has since been widely adopted for enhancing reasoning capabilities in LLMs.
Group relative advantage estimation method removes the need for value models entirely by evaluating responses relative to the average within a group of sampled responses.
This approach significantly reduces GPU memory usage and computational costs while maintaining competitive performance in downstream reasoning tasks.
Its effectiveness is further evidenced by the strong performance of modern reasoning models such as DeepSeek-R1 \citep{guo2025deepseek}, which highlights the effectiveness of group relative advantage estimation in balancing computational efficiency and performance robustness.
Several extensions have further refined this paradigm. Decoupled Clipping and Dynamic sAmpling Policy Optimization (DAPO) \citep{yu2025dapo} improves training on long CoT sequences through token-level gradient estimation and relaxed clipping strategies.
Dr. GRPO \citep{liu2025understanding} introduces an unbiased optimization method that enhances token efficiency, while Group Policy Gradient (GPG) \citep{chu2025gpg} further simplifies the learning process by removing surrogate losses and eliminating the reference model.
While these methods vary in implementation specifics, they all share a common foundation in the principle of group relative advantage estimation for effective LLMs post-training.
\begin{figure}[t]
    \centering
    \includegraphics[width=0.9\linewidth]{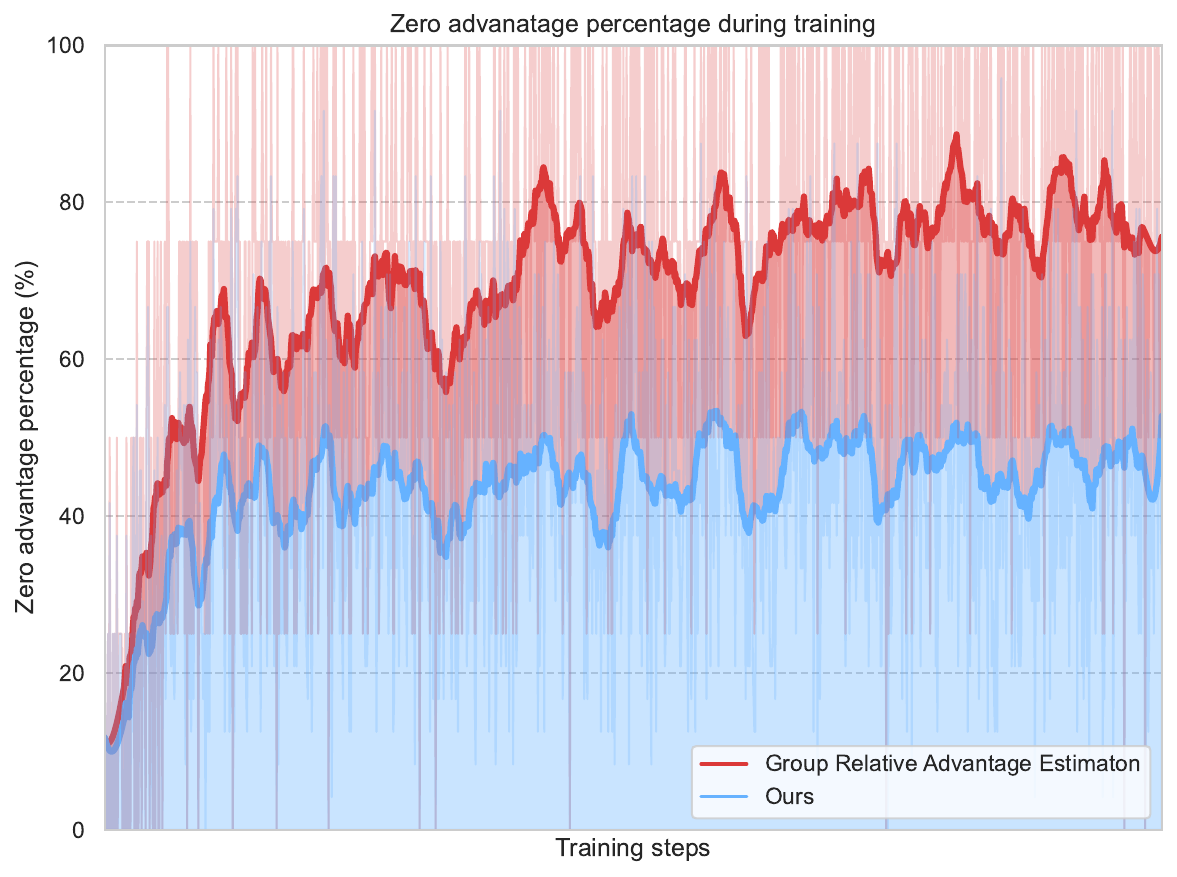}
    \caption{Statistical analysis of zero advantage proportion during training.}
    \label{fig:1}
\end{figure}

Although the above approaches have significantly advanced RL in the post-training stage and enhanced the reasoning capabilities of LLMs beyond what SFT can achieve, the practical limitations associated with the group relative advantage estimation method remain unresolved.
When the group relative advantage estimation method \citep{shao2024deepseekmath, chu2025gpg, liu2025understanding} is adopted, the advantage may approach zero when the rewards within a group exhibit low variance, resulting in zero gradient and, consequently, no parameter updates.
As shown in Figure~\ref{fig:1}, our statistical analysis of zero advantage proportion revealed that it is severe during training and approached even $100\%$ in the later training steps more frequently.
Conversely, when rewards among responses within the group vary significantly, the resulting advantages can exhibit high variance, potentially leading to unstable or unintended gradient ascent. Both scenarios deviate from the desired optimization trajectories.
In this work, to address the challenges of policy optimization from the group relative advantage estimation method mentioned above, we propose a novel RL algorithm Advantage-Augmented Policy Optimization (\sysname), which mitigates the issues we address by estimating the advantage with advantage margin.
Advantage margin is defined as the distance between the rewards of responses from the policy model and those of the responses from the reference model.
This approach incorporates a reference gradient into the original gradient, thus providing a reliable optimization signal that reflects the overall direction of improvement, even when the advantages approach zero.
Experiments on several representative mathematical reasoning benchmarks demonstrate the effectiveness and robustness of \sysname.

Our main contributions are as follows:
\begin{itemize}
    \item We delve into the optimization behavior of current RL algorithms adopting the group relative advantage estimation method, in the context of post-training LLMs with RL, with a particular emphasis on potential issues related to advantage estimation during the optimization.
    \item We propose Advantage-Augmented Policy Optimization (\sysname), which mitigates the issues of advantage estimation with advantage margin by taking a comparison with the reference model.
    \item Extensive experimental evaluation has demonstrated that \sysname achieves superior performance across different model series and mathematical reasoning benchmarks.
\end{itemize}

\section{Related Work}
\subsection{Reinforcement Learning for LLMs}
Early RL-based post-training methods \citep{christiano2017deep, song2024preference, ouyang2022training, ziegler2019fine} mainly relied on human-labeled preference data and reward models to evaluate responses.
PPO \citep{schulman2017proximal} estimates state values via a value model for advantage estimation while using a reward model to assign rewards.
DPO \citep{rafailov2023direct} uses paired preference data to encourage preferred responses and suppress undesired ones.
However, annotating preferences and training reward models are resource intensive.
Moreover, the scarcity of explicit reasoning data limits the enhancement of models’ reasoning capabilities via these RL methods.
DeepSeek-R1 is the first to report the use of RL to enable extended CoT generation and the emergence of the so-called "aha moment" \citep{guo2025deepseek}. Specifically, DeepSeek-R1 adopts GRPO \citep{shao2024deepseekmath}, which estimates advantages through in-group comparisons, thereby enhancing the feasibility of aligning models for reasoning tasks.
Yet, the limitations of existing advantage estimation remain underexplored.
In this work, we introduce \sysname, which redefines advantage estimation by incorporating advantage margin, effectively addressing the issues of the group relative advantage estimation method.
\subsection{Advantage estimation in RL}
Previous RL approaches \citep{schulman2017proximal, schulman2015trust, haarnoja2018soft} estimate the advantage using either Monte Carlo returns \citep{williams1992simple}, Temporal-Difference (TD) \citep{sutton1999policy} errors, or Generalized Advantage Estimation (GAE) \citep{schulman2015high}.
While these critic-based estimators form the backbone of algorithms such as PPO, maintaining a separate value network becomes computationally expensive when the policy is implemented as an LLM. 
Group relative advantage estimation method proposed by GRPO \citep{shao2024deepseekmath} computes the advantage by comparing each response's reward to the group mean, reducing the resources needed.
However, it introduces new optimization challenges, as discussed in the introduction section, that have yet to be investigated.
In this work, we propose a novel advantage estimation method that incorporates advantage margin and optimizes the cross-entropy loss to better enhance the reasoning capabilities of LLMs.

\section{Preliminary}
\label{preliminary}
This section provides a preliminary overview of group-based relative advantage estimation methods for RL-based training by briefly introducing GRPO \citep{shao2024deepseekmath} and GPG \citep{chu2025gpg}.
In the context of the GRPO algorithm, it circumvents the dependency on the value model which is commonly required in PPO \citep{schulman2017proximal} by estimating advantage within grouped samples.
It utilizes a rule-based reward system to score responses generated by LLMs.
Furthermore, GRPO retains the clipping strategy from PPO to prevent excessively large policy updates and leverages the reference model $\pi_{r e f}$ to compute the current Kullback-Leibler (KL) divergence, thereby ensuring the stability and integrity of the model during training.
Specifically, for each question $q$, GRPO samples a group of responses $O=\{o_{1},o_{2},\cdots,o_{G}\}$ from the old policy $\pi_{\theta_{\mathrm{old}}}$ and optimizes the policy model $\pi_\theta$ by minimizing the following objective:
\begin{equation}\label{eq:1}
\scalebox{0.61}{
$\displaystyle
\begin{aligned}
\mathcal{J}_{\mathrm{GRPO}}(\theta) & =\mathbb{E}_{(q, a) \sim \mathcal{D},\left\{o_i\right\}_{i=1}^G \sim \pi_{\theta_{\mathrm{old}}}(\cdot \mid q)} \frac{1}{G} \sum_{i=1}^G \frac{1}{\left|o_i\right|} \sum_{t=1}^{\left|o_i\right|}\\
&-\Bigg[\min \left[r_{i, t}(\theta)\hat{A}_{i,t}^{\mathrm{GRPO}}, \operatorname{clip}\left(r_{i, t}(\theta), 1-\varepsilon, 1+\varepsilon\right) \hat{A}_{i,t}^{\mathrm{GRPO}}\right]\\
&-\beta \mathbb{D}_{K L}\left[\pi_\theta| | \pi_{r e f}\right]\Bigg]
\end{aligned}
$}
\end{equation}
where $\varepsilon$ and $\beta$ are hyper-parameters to control the clip boundary and the KL divergence penalty coefficient, respectively. 
In this context, $\hat{A}_{i,t}^{\mathrm{GRPO}}$ represents the advantage, derived exclusively from the relative reward of the responses within the same group.
$r_{i, t}(\theta)$ represents the likelihood ratio between the current policy $\pi_{\theta}$ and the old policy $\pi_{\theta_{\mathrm{old}}}$.
Typically, the likelihood ratio $r_{i, t}(\theta)$ and the advantage $\hat{A}_{i,t}^{\mathrm{GRPO}}$ are calculated as:
\begin{equation}\label{eq:2}
\scalebox{0.67}{$\displaystyle
\begin{aligned}
r_{i, t}(\theta)=\frac{\pi_\theta\left(o_{i, t} \mid q, o_{i,<t}\right)}{\pi_{\theta_{\text {old }}}\left(o_{i, t} \mid q, o_{i,<t}\right)}, \quad \hat{A}_{i,t}^{\mathrm{GRPO}}=\frac{r_{i}-\operatorname{mean}(\{R_i\}_{i=1}^G)}{\operatorname{std}(\{R_i\}_{i=1}^G)}.
\end{aligned}
$}
\end{equation}

Recent work GPG \citep{chu2025gpg} proposes directly optimizing the original RL objective, yielding improved performance over GRPO.
While the advantage estimation method used in GPG is similar to that of GRPO, it further emphasizes the advantage of responses that are valid to gradient estimation.
Generally speaking, the core objective of GPG is to optimize the following objective:
\begin{equation}\label{eq:3}
\scalebox{0.68}{$\displaystyle
\begin{aligned}
\mathcal{J}_{\mathrm{GPG}}(\theta)&= \mathbb{E}_{(q, a) \sim \mathcal{D},\left\{o_i\right\}_{i=1}^G}\\
& {\Bigg[\frac{1}{\sum_{i=1}^G\left|o_i\right|} \sum_{i=1}^G \sum_{t=1}^{\left|o_i\right|}\left(-\log \pi_\theta\left(o_{i, t} \mid q, o_{i,<t}\right) \hat{A}_{i, t}^{\mathrm{GPG}}\right)\Bigg], }
\end{aligned}
$}
\end{equation}
where $\hat{A}_{i,t}^{\mathrm{GPG}}=\frac{r_{i}-\operatorname{mean}(\{R_i\}_{i=1}^G)}{F_{norm}}$ and $F_{norm}$ could be 1 or $\operatorname{std}(\{R_i\}_{i=1}^G)$.
However, the rewards of the responses within a group may exhibit low variance.
This suggests that $\hat{A}_{i,t}^{\mathrm{GPG}}$ in GPG encounters the same challenges as $\hat{A}_{i,t}^{\mathrm{GRPO}}$ in GRPO.

In this work, we propose a novel algorithm \sysname to address the issues caused by the group relative advantage estimation method in the policy optimization process.
We also provide an in-depth analysis of both the prevailing advantage estimation method adopted in \citep{shao2024deepseekmath, yu2025dapo, chu2025gpg} and our proposed \sysname.

\section{Advantage-Augmented Policy Optimization}
\label{aapo}

Inspired by the optimization behavior found in current RL algorithms \citep{shao2024deepseekmath, liu2025understanding, yu2025dapo, chu2025gpg}, our objective is to mitigate the issues that arise in the policy optimization process, overcoming the challenge where advantage estimation tends to approach zero or bad advantage estimation in the later steps of RL training.
Drawing further inspiration from the well-established Adam \citep{kingma2014adam} and the recent GPG algorithm \citep{chu2025gpg}, which optimizes the RL objective directly, thus avoiding the surrogate loss function, we propose a novel algorithm, Advantage-Augmented Policy Optimization (\sysname), which directly optimizes the cross-entropy (CE) loss enhanced by augmented advantage, which is driven by the advantage margin.
In contrast to previous approaches \citep{shao2024deepseekmath, chu2025gpg, liu2025understanding}, \sysname leverages advantage amplification by performing group-based sampling for both the policy model $\pi_\theta$ and the reference model $\pi_{r e f}$.
Specifically, we calculate the reward for each response generated by the policy model $\pi_\theta$, evaluate the relative advantage of each response within its group $G$, and compare these rewards $r_{\theta_i}$ with those rewards $r_{r e f_i}$ obtained from the reference model $\pi_{r e f}$. 
This method improves the effectiveness of the RL training step by preventing the advantage from approaching zero and exhibiting high variance.
Formally, \sysname optimizes the policy model $\pi_\theta$ by minimizing the following objective:
\begin{equation}\label{eq:4}
\scalebox{0.70}{$\displaystyle
\begin{aligned}
\mathcal{J}_{\mathrm{\sysname}}(\theta)&= \mathbb{E}_{(q, a) \sim \mathcal{D},\left\{o_i\right\}_{i=1}^G}\\
&{\Bigg[\frac{1}{\sum_{i=1}^G\left|o_i\right|} \sum_{i=1}^G \sum_{t=1}^{\left|o_i\right|}\left(-\log \pi_\theta\left(o_{i, t} \mid q, o_{i,<t}\right) \hat{A}_{i, t}^*\right)\Bigg], }
\end{aligned}
$}
\end{equation}
where $\hat{A}_{i,t}^*$ is computed using Equation~\eqref{eq:5} and $\operatorname{clip}$ operation is added to improve stability, which is discussed in Section~\ref{sec: ab}.
\begin{equation}\label{eq:5}
\scalebox{0.72}{$\displaystyle
\begin{aligned}
    \hat{A}_{i,t}^*=\frac{r_{\theta_i}-\operatorname{mean}(r_\theta)}{\operatorname{std}(r_\theta)} + \operatorname{clip}(\underbrace{r_{\theta_i} - r_{r e f_i}}_{\textbf{Advantage margin}}, \delta_{\mathrm{low}}, \delta_{\mathrm{high}}).
\end{aligned}
$}
\end{equation}

By Equation~\eqref{eq:4}, we can derive its gradient:
\begin{equation}\label{eq:6}
\scalebox{0.67}{$\displaystyle
\begin{aligned}
    \nabla_\theta \mathcal{J}_{\mathrm{\sysname}}(\theta)&=-\mathbb{E}_{(q, a) \sim \mathcal{D},\left\{o_i\right\}_{i=1}^G}\\
    &\Bigg[\frac{1}{\sum_{i=1}^G\left|o_i\right|} \sum_{i=1}^G \sum_{t=1}^{\left|o_i\right|}\hat{A}_{i, t}^* \cdot \nabla_\theta \log \pi_\theta\left(o_{i, t} \mid q, o_{i,<t}\right)\Bigg],
\end{aligned}
$}
\end{equation}
where the augmented advantage $\hat{A}_{i, t}^*$ functions as a constant coefficient that scales the gradient.

To provide theoretical guarantees for the training dynamics of \sysname, we explicitly analyze its stability and convergence properties.
We begin by formalizing the empirical loss function over a sampled group, which serves as the foundation for our theoretical derivation.\\
\textbf{Definition} \quad \textit{For a group $\mathcal{G}$ containing sampled responses $O=\{o_1,o_2,\cdots,o_\mathcal{G}\}$, the empirical \sysname loss is defined as $\mathcal{L}_{\mathcal{G}}(\theta)=\frac{1}{N_{\mathcal{G}}}\sum_{o \in \mathcal{G}}[-\log\pi_{\theta}(o)\hat{A}^{*}]$, where $\pi_\theta$ is the policy model, $N_{\mathcal{G}}=\sum_{o \in \mathcal{G}} |o|$ is the total number of tokens in the group.
We further define the expected objective as the expectation of the empirical loss over all possible groups $\mathcal{L}(\theta)=\mathbb{E}_{\mathcal{G}\sim\pi_{\theta}}[\mathcal{L}_{\mathcal{G}}(\theta)]$.}

First, we establish the stability of the \sysname training process. 
It is crucial to ensure that the introduction of the augmented advantage term does not lead to unbounded parameter updates.

\begin{theorem}
(Stability) \quad Since the rewards are bounded, the group standard deviation satisfies $0\leq\sigma_{min}\leq\sigma$, and the log-likelihood gradients are bounded as $||\nabla_{\theta}\log\pi_{\theta}(o)||\leq M$.
Then, each gradient step with learning rate $\eta_k$ satisfies $||\theta_{k+1}-\theta_k||\leq\eta_{k}MB$, where $B=\frac{R_{max}-R_{min}}{\sigma_{min}}+\operatorname{max}(|\delta_\mathrm{low}|, |\delta_\mathrm{high}|)$ is the uniform bound on the \sysname weights.
The expected objective is bounded from $\mathcal{L}(\theta)\geq-B\log|\mathcal{V}|$, where $|\mathcal{V}|$ is the vocabulary size.
Hence, \sysname training is stable: the objective cannot diverge to $-\infty$ and parameter updates are always finite.
Proof in Appendix~\ref{sec:proof}.
\end{theorem}

Beyond stability, we ensure that \sysname effectively minimizes the loss and locates a stationary point.
The following theorem guarantees the convergence of \sysname under standard stochastic approximation assumptions.

\begin{theorem}
(Convergence) \quad Assume that the stochastic gradient is unbiased and that the per-sample gradient has bounded second moment.
Let the step sizes satisfy the Robbins–Monro conditions $\eta_k>0$, $\sum_{k}\eta_k=\infty$, $\sum_{k}\eta_k^2<\infty$. 
\sysname converges to a stationary point of its expected objective $\liminf_{k \to \infty} \mathbb{E}\left[ \left\| \nabla \mathcal{L}(\theta_k) \right\|^2 \right] = 0$.
Moreover, if a constant step size $\eta<\frac{1}{BL_0}$ is used, where $L_0$ is the smoothness constant of $-\log\pi_{\theta}(o)$, then the iterates converge to a neighborhood of stationarity $\limsup_{K \to \infty} \frac{1}{K}\sum_{k=1}^K \mathbb{E}\!\left[ \left\| \nabla \mathcal{L}(\theta_k) \right\|^2 \right] \;\lesssim\; \mathcal{O}(\eta) + \mathcal{O}\!\left(\tfrac{1}{N_\mathcal{G}}\right)$.
Proof in Appendix~\ref{sec:proof}.
\end{theorem}

From the formulation of the augmented advantage $\hat{A}_{i, t}^*$, it is evident that as the policy is optimized, the reward $r_{\theta_i}$ of the responses sampled from the policy $\pi_\theta$ within a group increases.
Consequently, the distance between these rewards $r_{\theta_i}$ and those $r_{r e f_i}$ of the responses sampled from the reference model $\pi_{r e f}$ will also widen, since the parameters of the reference model remain frozen throughout the training process of \sysname.
In later steps of RL, the responses sampled from the policy tend to be of high quality, causing the relative advantages $\hat{A}_{i, t}^{\mathrm{GRPO}}$ within the group to approach zero.
If training continues using the original advantage estimation method in Equation~\eqref{eq:2}, the resulting gradients will approach zero, leading to significantly reduced training efficiency.
However, under the proposed method augmenting advantage with advantage margin in Equation~\eqref{eq:5}, even when the group relative advantage approaches zero, the rewards of the policy samples remain higher than those of the reference samples. 
This ensures a nonzero advantage $\hat{A}_{i, t}^*$ in \sysname as discussed in Section~\ref{how}, thereby maintaining informative gradients for continued policy optimization.

\section{Analysis of \sysname}
\label{analyss}
\subsection{Deep Analysis of Group Relative Advantage Estimation}
\label{analysis:1}
As shown in Equation~\eqref{eq:2}, current advantage estimation methods \citep{shao2024deepseekmath, yu2025dapo, liu2025understanding} that eliminate the dependency on a value model predominantly adopt this form of computation.
We now present a rigorous analysis into the underlying phenomena induced by this advantage estimation method. \textbf{Phenomenon 1}: What are the implications when all responses within a group are similarly good (or all of high quality)? \textbf{Phenomenon 2}: What are the implications when all responses within a group are similarly bad (or all of low quality)?

To rigorously address the aforementioned questions, we proceed with a systematic, step-by-step analysis.
For the sake of mathematical convenience in the proof, we limit our discussion to cases where all responses are similarly good in Phenomenon 1 and similarly bad in Phenomenon 2, respectively, as the proof for all of high quality and low quality responses follows the same structure.\\
\textbf{Phenomenon 1}\quad Considering that all responses are similarly good, which implies that the reward for each response is nearly the same, this indicates $\forall i,j \in \{1,2,3,\cdots,G\} \land i \neq j, \text{ } r_{i} \approx r_{j}$, their respective advantage, as estimated according to Equation~\eqref{eq:2}, can be expressed as following:
\begin{align}
    \hat{A}_{i, t}^{\mathrm{GRPO}}=\frac{r_{i}-\operatorname{mean}(\{R_i\}_{i=1}^G)}{\operatorname{std}(\{R_i\}_{i=1}^G)}\to0.\label{eq:7}
\end{align}

As expressed in Equation~\eqref{eq:7}, the advantage of each response approaches zero in this case.
For illustrative purposes, we use the loss function of GRPO \citep{yu2025dapo} as an example.
As mentioned in DAPO \citep{yu2025dapo}, removing the KL divergence in GRPO could further improve the optimization.
The KL divergence, summation and averaging operators in Equation~\eqref{eq:1} are omitted to facilitate clarity in this context, the gradient formula of GRPO is formally given by:
\begin{equation}\label{eq:8}
\scalebox{0.9}{$\displaystyle\nabla_\theta \mathcal{J}_{\mathrm{GRPO}}(\theta)=A_{i,t}^{\mathrm{GRPO}}\nabla_{\theta}\log\pi_{\theta}(o_{i,t}\mid q,o_{i,<t}).$}
\end{equation}

As illustrated in the computation of Equation~\eqref{eq:8}, when the advantage of each response approaches zero, the corresponding gradient also decreases to zero.
This implies that the gradient update for the policy becomes negligible for this training step, resulting in very low training efficiency.
Similar issues are observed in methods such as GRPO, DAPO, GPG \citep{chu2025gpg}, and Dr. GRPO \citep{liu2025understanding}.
Nonetheless, the observation that all responses are associated with similarly high reward does not unequivocally imply that the policy has been sufficiently optimized; it may alternatively reflect a high variance \citep{roelofs2019meta, yu2022understanding} in the current policy, suggesting that the policy has converged to a sub-optimum, performing well only on a narrow category of questions.
However, this phenomenon is frequently observed during the later steps of RL training if the advantage estimation method in GRPO is adopted.\\
\textbf{Phenomenon 2}\quad Similar to the scenario in Phenomenon 1 where all responses are similarly good, when all responses are similarly bad, the reward assigned to each response tends to be similar.
As a result, the relative advantage of each response approaches zero, leading to a zero gradient during policy updates, which is computed using Equation~\eqref{eq:8}.
Consequently, the efficiency of this training step becomes significantly low.
Such a phenomenon is frequently observed when the policy is confronted with input samples that exhibit inherently high complexity or ambiguous representation.\\
\textbf{Analysis Conclusion}\quad Based on our in-depth analysis of the two phenomena where generated responses are similarly good and similarly bad within a group, we observed that the gradient tends to approach zero.
This results in extremely low training efficiency during RL training. \\
\textbf{Above phenomena can be generalized}\quad When the rewards of responses within any given group are identical or highly similar, regardless of whether the responses are all good or all bad, the gradient approaches zero, rendering the gradient update nearly ineffective in training. 
This issue becomes particularly pronounced in the later steps of RL training, where generated responses are consistently of high quality, and the corresponding advantage approaches zero.
This indicates that the training efficiency progressively declines as RL training progresses.
Motivated by this insight, we propose an advantage-augmented RL algorithm \sysname to address the aforementioned phenomena.

\begin{table*}[t]
  \centering
  \small
  \setlength{\tabcolsep}{2pt}
  \begin{adjustbox}{width=\textwidth, center} 
  \begin{tabular}{lcccccc>{\columncolor{mygray}}c}
    \toprule
    Model&Training Samples&AIME24&MATH-500&AMC23&Minerva&OlympiadBench&Avg. \\
    \midrule
    \multicolumn{7}{c}{\textit{Llama 1B Models}} \\
    \midrule
    Llama-3.2-1B-Instruct$^{\dagger}$ & -- & 0.0 & 11.8 & 2.5 & 1.8 & 3.7 & 4.0 \\
    +GRPO$^{\dagger}$ \citep{shao2024deepseekmath} & 8,523 & 0.0 & 19.4 & 12.5 & 3.7 & 4.3 & 8.0 \\
    +GPG$^{\dagger}$ \citep{chu2025gpg} & 8,523 & 0.0 & 21.2 & 17.5 & 1.8 & 4.7 & \underline{9.0} \\
    \rowcolor{gray!15}
    +\sysname \ (Ours) & 8,523 & 10.0 & 25.0 & 12.5 & 4.0 & 6.1 & \textbf{11.5} \\
    \midrule
    \multicolumn{7}{c}{\textit{Llama 3B Models}} \\
    \midrule
    Llama-3.2-3B-Instruct$^{\dagger}$ & -- & 3.3 & 28.6 & 7.5 & 3.7 & 7.6 & 10.1 \\
    +GRPO$^{\dagger}$ \citep{shao2024deepseekmath} & 8,523 & 0.0 & 32.8 & 22.5 & 8.1 & 7.7 & 14.2 \\
    +GPG$^{\dagger}$ \citep{chu2025gpg} & 8,523 & 6.7 & 40.0 & 15.0 & 8.8 & 11.6 & \underline{16.2} \\
    \rowcolor{gray!15}
    +\sysname (Ours) & 8,523 & 6.7 & 43.8 & 22.5 & 9.6 & 11.4 & \textbf{18.8} \\
    \midrule
    \multicolumn{7}{c}{\textit{Qwen 1.5B Models}} \\
    \midrule
    {\small DeepSeek-R1-Distill-Qwen-1.5B$^{\dagger}$} & -- & 33.3 & 84.4 & 70.0 & 30.9 & 50.8 & 53.9 \\
    {\small GRPO-1.5B$^{\dagger}$} \citep{dang2025reinforcement} & 7,000 & 26.7 & 86.2 & 82.5 & 27.6 & 52.6 & 55.2 \\
    {\small GPG-1.5B$^{\dagger}$ \citep{chu2025gpg}} & 7,000 & 36.7 & 83.4 & 75.0 & 29.8 & 53.2 & 55.6 \\
    {\small Still-3-1.5B-Preview}$^{\dagger}$ \citep{chen2025empirical} & 30K & 40.0 & 85.5 & 72.5 & 30.5 & 53.9 & \underline{56.5} \\
    \rowcolor{gray!15}
    {\small \sysname-1.5B (Ours)} & 7,000 & 33.3 & 86.0 & 80.0 & 30.9 & 53.3 & \textbf{56.7} \\
    \midrule
    \multicolumn{7}{c}{\textit{Qwen 7B Models}} \\
    \midrule
    {\small Qwen2.5-Math-7B$^{\dagger}$ \citep{yang2024qwen2math}} & -- & 6.7 & 56.2 & 47.5 & 14.0 & 23.4 & 39.6 \\
    {\small SimpleRL-Zero-7B$^{\dagger}$ \citep{Qwne2.5-7b-simplerl}}& 8,523 & 30.0 & 77.4 & 57.5 & 30.5 & 38.1 & 46.7 \\
    {\small GPG-7B$^{\dagger}$ \citep{chu2025gpg}} & 8,523 & 23.3 & 80.2 & 55.0 & 36.0 & 42.8 & 47.5 \\
    {\small OpenReasoner-Zero-7B$^{\dagger}$ \citep{hu2025open}} & 57K & 20.0 & 80.8 & 65.0 & 29.4 & 46.2 & 48.3 \\
    {\small Eurus-2-7B-PRIME$^{\dagger}$ \citep{cui2025process}} & 230K + 150K & 16.7 & 81.8 & 65.0 & 37.5 & 44.6 & 49.1 \\
    {\small Oat-Zero-7B$^{\dagger}$ \citep{liu2025understanding}} & 8,523 & 30.0 & 81.2 & 65.0 & 34.9 & 43.4 & \underline{50.3} \\
    \rowcolor{gray!15}
    {\small \sysname-7B (Ours)} & 8,523 & 30.0 & 82.4 & 70.0 & 35.3 & 44.3 & \textbf{52.4} \\
    \bottomrule
  \end{tabular}
  \end{adjustbox}
  \caption{Zero-shot pass@1 performance on mathematical reasoning benchmarks. $^{\dagger}$ represents reproduced results with our best effort under the same settings. \textbf{Bold} and \underline{Underline} indicate the best and the second-best performance in the corresponding category, respectively.}
\label{tab:1}
\end{table*}
\subsection{Understanding the effectiveness of \sysname}
\label{how}
As discussed in Section~\ref{analysis:1}, the previous advantage estimation method (as expressed in Equation~\eqref{eq:2}) in RL algorithms \citep{shao2024deepseekmath, chu2025gpg, liu2025understanding} can lead to the advantage value approaching zero, which in turn causes the magnitude of gradient updates to diminish accordingly.
In this section, we provide a comprehensive analysis of why our proposed advantage-augmented method can effectively mitigate these issues.\\
\textbf{Analysis 1}\quad Consider a general situation, which naturally encompasses the two phenomena in Section~\ref{analysis:1}.
In the later steps of RL training, once the policy has already acquired relatively easier-to-learn features, it often struggles to learn more complex ones.
During this phase, when the policy $\pi_\theta$ samples a group of responses, the reward associated with each sample tends to be similar.
Consequently, the relative advantage computed according to Equation~\eqref{eq:2} approaches zero.
To address this issue, we propose \sysname, which estimates the advantage $\hat{A}_{i, t}^*$ with advantage margin following Equation~\eqref{eq:5}. 
Since the capability of the reference model $\pi_{r e f}$ remains unchanged while the policy model $\pi_\theta$ improves progressively throughout \sysname training, the quality of responses generated by the policy model $\pi_\theta$ exceeds that of the reference model $\pi_{r e f}$ over time.
By measuring the distance between the rewards of two groups of responses $O_{\theta}=\{o_{\theta_1}, o_{\theta_2}, \cdots, o_{\theta_G}\}$ and $O_{r e f}=\{o_{r e f_{1}}, o_{r e f_{2}}, \cdots, o_{r e f_{G}}\}$ from $\pi_{r e f}$ generated by the policy model $\pi_\theta$ and the reference model $\pi_{r e f}$, respectively, we can calculate the augmented advantages of each response in $O_\theta$ via Equation~\eqref{eq:5}.
This prevents the advantages from approaching zero and ensures that the gradients used to update the policy remain informative and effective.
This analysis can be generalized to any situation where the rewards of the responses in the group are similar, whether good or not.\\
\textbf{Analysis 2} \quad Group relative advantage estimation (Equation~\eqref{eq:2}) can be problematic in the presence of two types of asymmetric response distributions.
In one case, where most responses from the policy $\pi_\theta$ are high quality except for a single outlier, the relatively worse response may receive a negative advantage and contribute an opposing gradient, increasing variance.
Although this response may still be correct under multi-dimensional reward rules, such as both format and correctness, it is penalized due to its format.
This misalignment between reward attribution and the true value of a response increases the risk of reward hacking\citep{everitt2021reward, pan2022effects}.
In the opposite case, where most responses are low quality and one is relatively better, the better one receives an excessively high advantage despite possibly low absolute quality, leading to biased updates and suboptimal convergence.
\sysname augments advantage estimation with advantage margin, addresses both issues: it boosts underappreciated yet valuable responses and suppresses misleading ones that exhibit a high estimated advantage.
As a result, \sysname reduces variance and the risk of reward hacking, and improves the optimization stability of the policy.\\
\textbf{Analysis 3} \quad When policy optimization with \sysname reaches a global optimum, which means $\forall i,j \in \{1,2,3,\cdots,G\} \land i \neq j, \text{ } r_{i} \approx r_{j}$, the objective in Equation~\eqref{eq:6}, which consists of the CE loss and the augmented advantage, exhibits a specific and predictable behavior as detailed in our derivation below (for clarity and notational convenience, we omit the expectation operator).
The optimization exhibits slight oscillations near the optimum, avoiding large gradient updates.
\begin{equation}\notag
\scalebox{0.66}{$\displaystyle
\begin{aligned}
\mathcal{J}_{\mathrm{\sysname}}(\theta)&\approx\frac{1}{\sum_{i=1}^G\left|o_i\right|} \sum_{i=1}^G \sum_{t=1}^{\left|o_i\right|}\\
&\Bigg[-\log \pi_\theta\left(o_{i, t} \mid q, o_{i,<t}\right) \cdot \operatorname{clip}(r_{\theta_i} - r_{r e f_i}, \delta_{\mathrm{low}}, \delta_{\mathrm{high}})\Bigg]\\
 &\leq \delta_{\mathrm{high}} \cdot {\frac{1}{\sum_{i=1}^G\left|o_i\right|} \sum_{i=1}^G \sum_{t=1}^{\left|o_i\right|}\left(-\log \pi_\theta\left(o_{i, t} \mid q, o_{i,<t}\right)\right)}
\end{aligned}
$}
\end{equation}

\section{Experiments}
\label{experiments}
\subsection{Experimental setup}
Most recent RL algorithms \citep{shao2024deepseekmath, yu2025dapo, hu2025reinforce++, liu2025understanding} struggle to make LLMs perform better at solving mathematical problems, which requires the model to think in the CoT \citep{wei2022chain} format before deciding the final answer.
We choose open-rs \citep{dang2025reinforcement} as our training dataset for DeepSeek-R1-Distill-Qwen-1.5B base model, because the data in this dataset cover various types and difficulty levels of mathematical problems that are highly representative.
To further provide a fair and rigorous evaluation of the effectiveness of our proposed \sysname, Qwen2.5-Math-7B model is chosen as the base model and subsequently trained on more challenging simplelr\_qwen\_level3to5 dataset \citep{Qwne2.5-7b-simplerl}.
In addition, we also validate the effectiveness of \sysname on Llama series models.

In our experiment setting, we set the clip parameters $\delta_{\mathrm{low}}$ and $\delta_{\mathrm{high}}$ to be -0.2 and 0.28, respectively.
We train all base models under the \sysname following the training process depicted in Algorithm \ref{alg:aapo}. 
All rule-based reward functions adopted in our experiments are simple.
More training details about rule-based reward functions and the system prompt are provided in Appendix \ref{sec: train}.
To evaluate the extent to which our proposed \sysname algorithm can enhance the reasoning capabilities of the model, we select AIME24 (30 questions), MATH-500 (500 questions) \citep{hendrycks2021measuring, lightman2023let}, AMC23 (40 questions), Minerva (272 questions) \citep{lewkowycz2022solving}, and OlympiadBench (674 questions) \citep{huang2024olympicarena} as evaluation benchmarks. 
Our evaluation framework utilizes a well-established and community-vetted codebase, maintaining consistency with widely adopted implementations.

\subsection{Results}
\label{sec: result}
As shown in Table \ref{tab:1}, taking model DeepSeek-R1-Distill-Qwen-1.5B as our base model, the application of our proposed \sysname enables the base model to achieve the SOTA performance on Minerva, the second-best performance on MATH-500, AMC23 and OlympiadBench.
When averaging scores across all benchmarks, the resulting \sysname-1.5B model achieves an overall SOTA performance.
By direct comparison under the same training data, \sysname-1.5B achieves improvements of \textbf{2.7\%} and \textbf{2.0\%} over GRPO-1.5B and GPG-1.5B, respectively.
It is worth noting that our \sysname-1.5B achieves performance superior to Still-3-1.5B-Preview \citep{Still-3-1.5b-preview}, despite the fact that Still-3-1.5B-Preview benefits from a larger training dataset that consists of long CoT reasoning data distilled from the DeepSeek-R1 \citep{guo2025deepseek} model and performs RL training with specified reward strategies.
After employing our proposed \sysname on the Qwen2.5-Math-7B model, \sysname-7B achieves the overall SOTA performance compared to other methods.
\sysname-7B achieves improvements of \textbf{12.2\%}, \textbf{10.3\%} over SimpleRL-Zero-7B \citep{Qwne2.5-7b-simplerl}, GPG-7B \citep{chu2025gpg}, respectively, under identical training data.
Furthermore, \sysname-7B also outperforms other models such as Eurus-2-7B-PRIME \citep{cui2025process}, OpenReasoner-Zero-7B \citep{hu2025open}, despite the fact that these baselines are trained with more high-quality data or data distilled from the DeepSeek-R1 model.
This suggests that the effectiveness can be mainly attributed to the design of our proposed \sysname rather than to the scale or quality of the training data.
\sysname also demonstrates superior performance compared to GRPO and GPG on Llama series models under the same training and evaluation settings as reported in Table \ref{tab:1}.
Compared to GRPO and GPG, \sysname achieves absolute improvements of \textbf{3.5\%}, \textbf{2.5\%}, and \textbf{4.6\%}, \textbf{2.6\%} on Llama 1B and 3B models, respectively. 
More experimental results can be found in Appendix~\ref{appendix: more}.
\subsection{Ablation study}
\label{sec: ab}
\textbf{Clip operation}\quad To investigate the contribution of the clip operation to \sysname, we conduct additional ablation studies by removing the clip operation on both the 1.5B and 7B models.
The results presented in Table \ref{tab:2} indicate that the performance of the optimized model without clip is inferior to that with the clip, suggesting that incorporating the clip operation effectively contributes to further improvements in optimization results.
As illustrated in Appendix Figure~\ref{fig:clip}, the optimization process becomes more stable with the incorporation of the clip operation compared to the optimization process without it.
The resulting \sysname-7B exhibits better performance on the benchmarks when the clip operation is adopted.
\begin{table}[ht]
  \centering
  \setlength{\tabcolsep}{2pt}
  \begin{adjustbox}{width=0.48\textwidth, center} 
  \begin{tabular}{lccccc>{\columncolor{mygray}}c}
    \toprule
    Model&AIME24&MATH-500&AMC23&Minerva&OlympiadBench&Avg. \\
    \midrule
    {\small \sysname-1.5B} & 33.3 & 86.0 & 80.0 & 30.9 & 53.3 & \textbf{56.7} \\
    \midrule
    {\small \sysname-1.5B \textit{w/o} clip} & 33.3 & 85.0 & 82.5 & 29.0 & 53.3 & 56.6 \\
    \midrule
    {\small \sysname-7B} & 30.0 & 82.4 & 70.0 & 35.3 & 44.3 & \textbf{52.4}\\
    \midrule
    {\small \sysname-7B \textit{w/o} clip} & 30.0 & 79.6 & 70.0 & 34.9 & 42.5 & 51.2 \\
    \bottomrule
  \end{tabular}
  \end{adjustbox}
  \caption{Ablation study results. \textit{w/o} indicates without clip operation on the advantage margin. Zero-shot pass@1 performance on different benchmarks.}
\label{tab:2}
\end{table}\\
\textbf{Choice of the reference model} \quad Since the \sysname requires using the reference model to generate responses and obtain rewards $r_{ref_i}$, it is necessary to investigate the selection of the reference model.
We further explored two scenarios: using the base model (initial model) as the reference model and replacing the reference model with the policy model during training.
For replacing reference model, the reference model was continuously updated by replacing it with the current policy model, we update the reference model every 20 steps and 100 steps in \sysname-1.5B and \sysname-7B training process, respectively. 
Experimental results indicate that using the base model as the reference model yields the best performance. The experimental results are in Table~\ref{tab:4}.
Keeping the reference model as the initial model yields superior results because \sysname's core mechanism relies on the advantage margin in Equation~\eqref{eq:5}.
Updating the reference model during training reduces the value of advantage margin, making \sysname less effective.
\begin{table}[hb]
  \centering
  \setlength{\tabcolsep}{2pt}
  \begin{adjustbox}{width=0.48\textwidth, center} 
  \begin{tabular}{lccccc>{\columncolor{mygray}}c}
    \toprule
    Model&AIME24&MATH-500&AMC23&Minerva&OlympiadBench&Avg. \\
    \midrule
    {\small \sysname-1.5B} & 33.3 & 86.0 & 80.0 & 30.9 & 53.3 & \textbf{56.7} \\
    {\small \sysname-1.5B \textit{w} reference update} & 36.7 & 84.2 & 75.0 & 27.9 & 53.2 & 55.4 \\
    \midrule
    {\small \sysname-7B} & 30.0 & 82.4 & 70.0 & 35.3 & 44.3 & \textbf{52.4} \\
    {\small \sysname-7B \textit{w} reference update} & 26.7 & 80.6 & 70.0 & 33.8 & 44.0 & 51.0 \\
    \bottomrule
  \end{tabular}
  \end{adjustbox}
  \caption{Ablation on the choice of the reference model.}
\label{tab:4}
\end{table}
\begin{figure*}[t]
    \centering
    \begin{subfigure}[b]{0.32\textwidth}
        \centering
        \includegraphics[width=\textwidth]{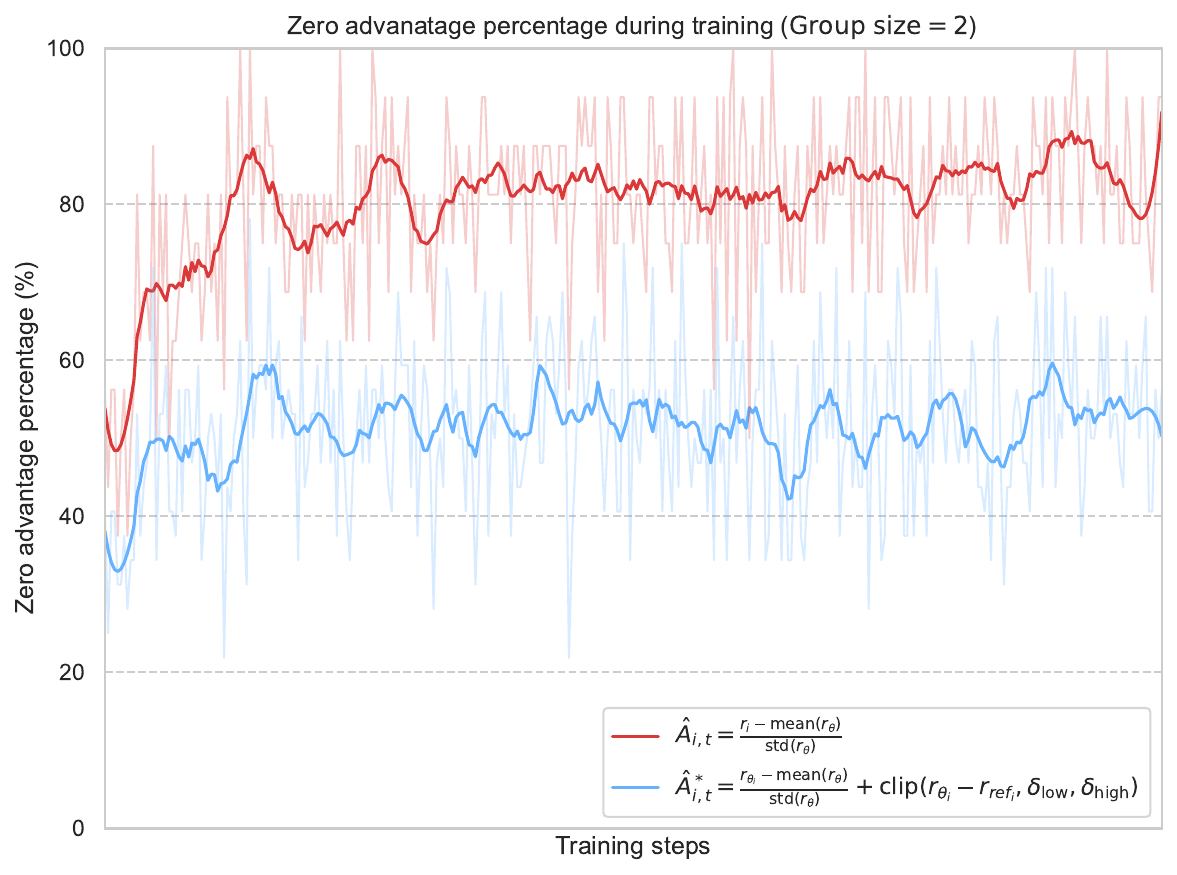}
    \end{subfigure}
    \begin{subfigure}[b]{0.32\textwidth}
        \centering
        \includegraphics[width=\textwidth]{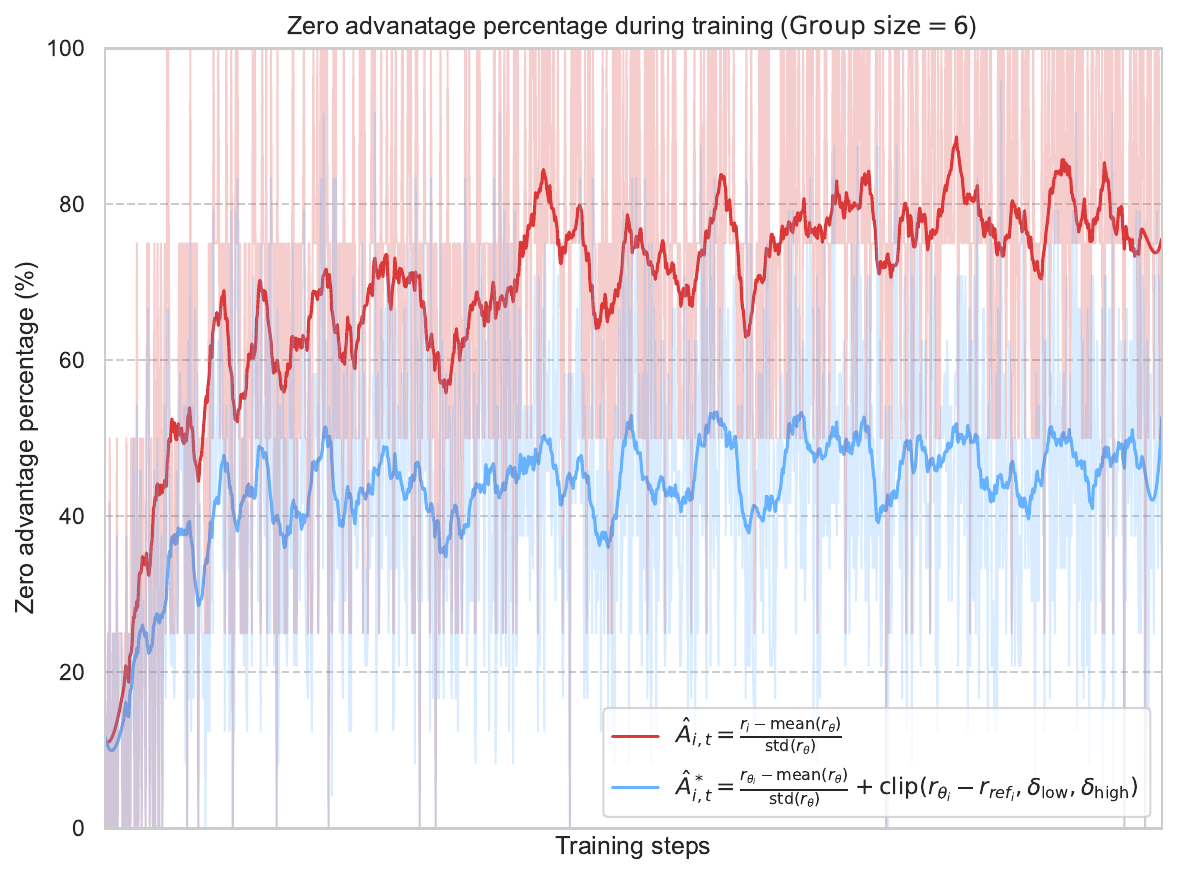}
    \end{subfigure}
    \begin{subfigure}[b]{0.32\textwidth}
        \centering
        \includegraphics[width=\textwidth]{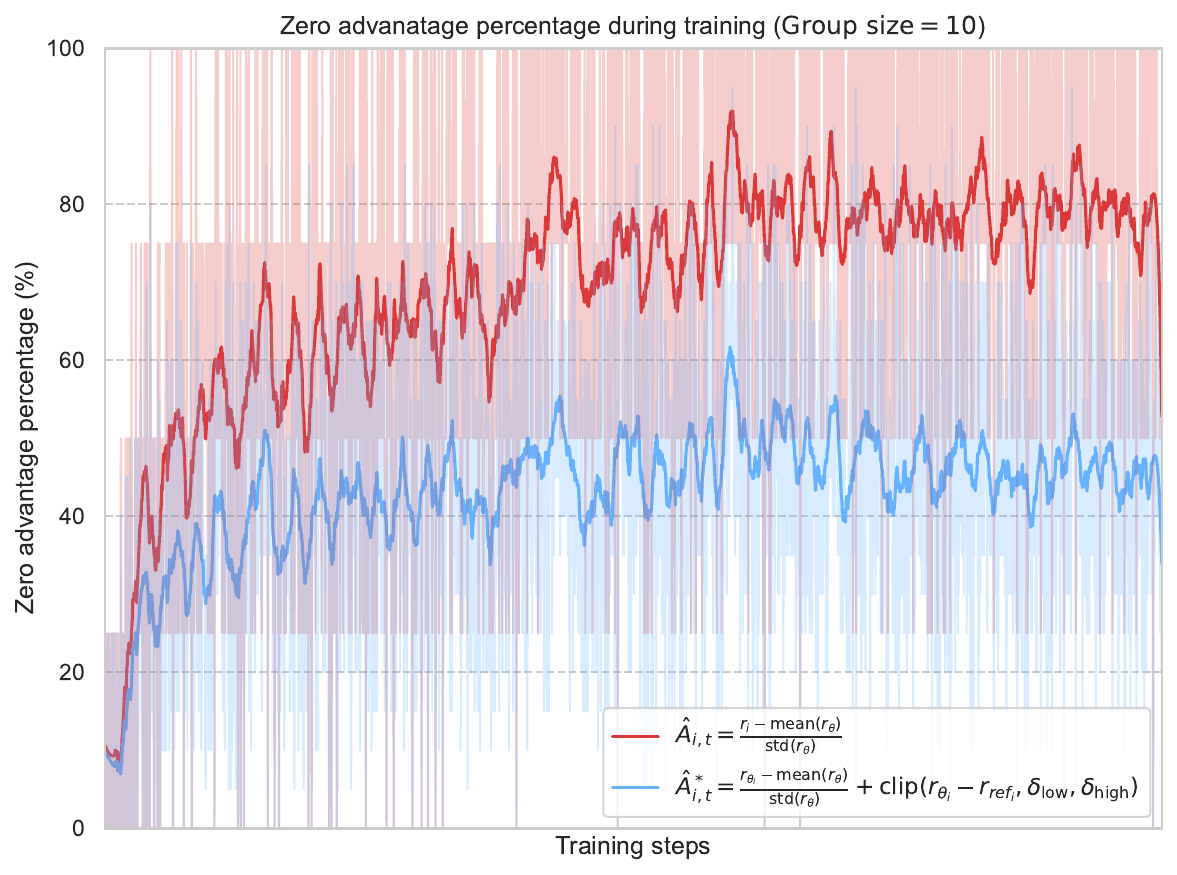}
    \end{subfigure}
    \caption{Training dynamics of the proportion of zero advantage with varying group size. We calculated the percentage of zero advantage under vanilla $A$ and our $A^{*}$ across different group size, presenting the average zero advantage percentage per device.}
    \label{fig:advantage}
\end{figure*}
\subsection{Training dynamics of advantage}
To further illustrate how our method \sysname mitigates the zero advantage phenomenon compared to the group relative advantage estimation method, we present the proportion of zero advantage per device during training across different group sizes. 
As shown in Figure~\ref{fig:advantage}, as the number of training steps increases, the vanilla group relative advantage estimation method $A$ (calculated by Equation~\eqref{eq:2}) exhibits a higher proportion of zero advantage per device.
In the later steps of training, the proportion of zero advantage increases significantly, and the frequency of all advantages being zero also increases markedly, leading to zero gradients and consequently reducing training efficiency.
However, during training with \sysname, regardless of the group size, no instance of all responses having all zero advantage occurred.
These training dynamics demonstrate that our $A^{*}$ (calculated by Equation~\eqref{eq:5}) effectively mitigates the occurrence of all zero advantage, transforming what would otherwise be ineffective gradient updates into meaningful ones.
\subsection{Computational efficiency}
Since \sysname requires sampling on the reference model, it introduces additional training overhead. 
We measured the training time and GPU memory usage per step using \sysname compared to GRPO on the same computational device.
The additional experimental information are shown in Table~\ref{tab:compute}.
\begin{table}[b]
  \centering
  \small
  \setlength{\tabcolsep}{22pt}
  \begin{adjustbox}{width=.48\textwidth} 
  \begin{tabular}{lcc}
    \toprule
    Method & Memory & Time\\
    \midrule
    \multicolumn{3}{c}{\textit{1.5B Model}} \\
    \midrule
    GRPO & 52.2GiB & 567s \\
    \sysname & 52.2GiB & 826s \\
    \midrule
    \multicolumn{3}{c}{\textit{7B Model}} \\
    \midrule
    GRPO & 78.7GiB & 299s \\
    \sysname & 78.9GiB & 384s \\
    \bottomrule
  \end{tabular}
  \end{adjustbox}
  \caption{Extra experimental information about computational efficiency.}
\label{tab:compute}
\end{table}
When training 1.5B model, \sysname and GRPO both use 52.2GiB GPU memory per device, \sysname takes 826s per step and GRPO takes 567s per step. 
When training 7B model, \sysname use 78.9GiB GPU memory per device and GRPO use 78.7GiB GPU memory per device, \sysname takes 384s per step and GRPO takes 299s per step.
It is clear that \sysname does not introduce extra GPU memory overhead and remains on par with GRPO.
The reason for why 1.5B model takes longer training time per step is that 1.5B model samples responses with much longer length (i.e., average\_response\_length $\approx$ 3000 for 1.5B model and average\_response\_length $\approx$ 700 for 7B model during training).
Since we use vanilla inference method (i.e., model.generate()) to generate responses, the extra time consumption could be significantly reduced using an advanced inference engine (e.g., VLLM) during training.

\subsection{More experimental results}
Experimental results about Out-of-Domain performance, more training dynamics of \sysname, and more ablation studies can be found in Appendix~\ref{appendix: more}.

\section{Conclusion}
\label{conclusion}
In this paper, we conduct an in-depth analysis of the limitations inherent in the group relative advantage estimation method used by mainstream RL algorithms, such as GRPO, which would lead to optimization issues such as zero gradient and gradient ascent. 
To address these issues, we propose a novel RL algorithm Advantage-Augmented Policy Optimization (\sysname).
By augmenting the group relative advantage estimation method with advantage margin, our method effectively improves policy optimization performance in experimental benchmarks.
Experimental results across several mathematical reasoning benchmarks and model series demonstrate that \sysname achieves the overall superior performance.

\section*{Limitations}
Our proposed \sysname can effectively mitigate the phenomenon that the estimated advantage approaches zero, but \sysname cannot eliminate this phenomenon, leaving this problem for further research.
Another limitation of our work lies in the long training time required, since \sysname depends on sampling responses from the reference model. 
However, this could be optimized by using advanced inference library (e.g. vLLM~\cite{kwon2023efficient}) to generate all responses.

\bibliography{main}

@article{shao2024deepseekmath,
  title={Deepseekmath: Pushing the limits of mathematical reasoning in open language models},
  author={Shao, Zhihong and Wang, Peiyi and Zhu, Qihao and Xu, Runxin and Song, Junxiao and Bi, Xiao and Zhang, Haowei and Zhang, Mingchuan and Li, YK and Wu, Y and others},
  journal={arXiv preprint arXiv:2402.03300},
  year={2024}
}

@article{hu2025reinforce++,
  title={REINFORCE++: A Simple and Efficient Approach for Aligning Large Language Models},
  author={Hu, Jian},
  journal={arXiv preprint arXiv:2501.03262},
  year={2025}
}

@article{rafailov2023direct,
  title={Direct preference optimization: Your language model is secretly a reward model},
  author={Rafailov, Rafael and Sharma, Archit and Mitchell, Eric and Manning, Christopher D and Ermon, Stefano and Finn, Chelsea},
  journal={Advances in Neural Information Processing Systems},
  year={2023}
}

@article{christiano2017deep,
  title={Deep reinforcement learning from human preferences},
  author={Christiano, Paul F and Leike, Jan and Brown, Tom and Martic, Miljan and Legg, Shane and Amodei, Dario},
  journal={Advances in Neural Information Processing Systems},
  year={2017}
}

@inproceedings{lightman2023let,
  title={Let's verify step by step},
  author={Lightman, Hunter and Kosaraju, Vineet and Burda, Yuri and Edwards, Harrison and Baker, Bowen and Lee, Teddy and Leike, Jan and Schulman, John and Sutskever, Ilya and Cobbe, Karl},
  booktitle={The Twelfth International Conference on Learning Representations},
  year={2024}
}

@inproceedings{song2024preference,
  title={Preference ranking optimization for human alignment},
  author={Song, Feifan and Yu, Bowen and Li, Minghao and Yu, Haiyang and Huang, Fei and Li, Yongbin and Wang, Houfeng},
  booktitle={Proceedings of the AAAI Conference on Artificial Intelligence},
  year={2024}
}

@article{chowdhery2023palm,
  title={Palm: Scaling language modeling with pathways},
  author={Chowdhery, Aakanksha and Narang, Sharan and Devlin, Jacob and Bosma, Maarten and Mishra, Gaurav and Roberts, Adam and Barham, Paul and Chung, Hyung Won and Sutton, Charles and Gehrmann, Sebastian and others},
  journal={Journal of Machine Learning Research},
  year={2023}
}

@article{brown2020language,
  title={Language models are few-shot learners},
  author={Brown, Tom and Mann, Benjamin and Ryder, Nick and Subbiah, Melanie and Kaplan, Jared D and Dhariwal, Prafulla and Neelakantan, Arvind and Shyam, Pranav and Sastry, Girish and Askell, Amanda and others},
  journal={Advances in Neural Information Processing Systems},
  year={2020}
}

@article{radford2018improving,
  title={Improving language understanding by generative pre-training},
  author={Radford, Alec and Narasimhan, Karthik and Salimans, Tim and Sutskever, Ilya and others},
  year={2018},
  publisher={San Francisco, CA, USA}
}

@article{bommasani2021opportunities,
  title={On the opportunities and risks of foundation models},
  author={Bommasani, Rishi and Hudson, Drew A and Adeli, Ehsan and Altman, Russ and Arora, Simran and von Arx, Sydney and Bernstein, Michael S and Bohg, Jeannette and Bosselut, Antoine and Brunskill, Emma and others},
  journal={arXiv preprint arXiv:2108.07258},
  year={2021}
}

@article{liu2023pre,
  title={Pre-train, prompt, and predict: A systematic survey of prompting methods in natural language processing},
  author={Liu, Pengfei and Yuan, Weizhe and Fu, Jinlan and Jiang, Zhengbao and Hayashi, Hiroaki and Neubig, Graham},
  journal={ACM computing surveys},
  year={2023},
  publisher={ACM New York, NY}
}

@article{wei2022chain,
  title={Chain-of-thought prompting elicits reasoning in large language models},
  author={Wei, Jason and Wang, Xuezhi and Schuurmans, Dale and Bosma, Maarten and Xia, Fei and Chi, Ed and Le, Quoc V and Zhou, Denny and others},
  journal={Advances in Neural Information Processing Systems},
  year={2022}
}

@Article{guo2025deepseek,
author={Guo, Daya
and Yang, Dejian
and Zhang, Haowei
and Song, Junxiao
and Wang, Peiyi
and Zhu, Qihao
and Xu, Runxin
and Zhang, Ruoyu
and Ma, Shirong
and Bi, Xiao
and Zhang, Xiaokang
and Yu, Xingkai
and Wu, Yu
and Wu, Z. F.
and Gou, Zhibin
and Shao, Zhihong
and Li, Zhuoshu
and Gao, Ziyi
and Liu, Aixin
and Xue, Bing
and Wang, Bingxuan
and Wu, Bochao
and Feng, Bei
and Lu, Chengda
and Zhao, Chenggang
and Deng, Chengqi
and Ruan, Chong
and Dai, Damai
and Chen, Deli
and Ji, Dongjie
and Li, Erhang
and Lin, Fangyun
and Dai, Fucong
and Luo, Fuli
and Hao, Guangbo
and Chen, Guanting
and Li, Guowei
and Zhang, H.
and Xu, Hanwei
and Ding, Honghui
and Gao, Huazuo
and Qu, Hui
and Li, Hui
and Guo, Jianzhong
and Li, Jiashi
and Chen, Jingchang
and Yuan, Jingyang
and Tu, Jinhao
and Qiu, Junjie
and Li, Junlong
and Cai, J. L.
and Ni, Jiaqi
and Liang, Jian
and Chen, Jin
and Dong, Kai
and Hu, Kai
and You, Kaichao
and Gao, Kaige
and Guan, Kang
and Huang, Kexin
and Yu, Kuai
and Wang, Lean
and Zhang, Lecong
and Zhao, Liang
and Wang, Litong
and Zhang, Liyue
and Xu, Lei
and Xia, Leyi
and Zhang, Mingchuan
and Zhang, Minghua
and Tang, Minghui
and Zhou, Mingxu
and Li, Meng
and Wang, Miaojun
and Li, Mingming
and Tian, Ning
and Huang, Panpan
and Zhang, Peng
and Wang, Qiancheng
and Chen, Qinyu
and Du, Qiushi
and Ge, Ruiqi
and Zhang, Ruisong
and Pan, Ruizhe
and Wang, Runji
and Chen, R. J.
and Jin, R. L.
and Chen, Ruyi
and Lu, Shanghao
and Zhou, Shangyan
and Chen, Shanhuang
and Ye, Shengfeng
and Wang, Shiyu
and Yu, Shuiping
and Zhou, Shunfeng
and Pan, Shuting
and Li, S. S.
and Zhou, Shuang
and Wu, Shaoqing
and Yun, Tao
and Pei, Tian
and Sun, Tianyu
and Wang, T.
and Zeng, Wangding
and Liu, Wen
and Liang, Wenfeng
and Gao, Wenjun
and Yu, Wenqin
and Zhang, Wentao
and Xiao, W. L.
and An, Wei
and Liu, Xiaodong
and Wang, Xiaohan
and Chen, Xiaokang
and Nie, Xiaotao
and Cheng, Xin
and Liu, Xin
and Xie, Xin
and Liu, Xingchao
and Yang, Xinyu
and Li, Xinyuan
and Su, Xuecheng
and Lin, Xuheng
and Li, X. Q.
and Jin, Xiangyue
and Shen, Xiaojin
and Chen, Xiaosha
and Sun, Xiaowen
and Wang, Xiaoxiang
and Song, Xinnan
and Zhou, Xinyi
and Wang, Xianzu
and Shan, Xinxia
and Li, Y. K.
and Wang, Y. Q.
and Wei, Y. X.
and Zhang, Yang
and Xu, Yanhong
and Li, Yao
and Zhao, Yao
and Sun, Yaofeng
and Wang, Yaohui
and Yu, Yi
and Zhang, Yichao
and Shi, Yifan
and Xiong, Yiliang
and He, Ying
and Piao, Yishi
and Wang, Yisong
and Tan, Yixuan
and Ma, Yiyang
and Liu, Yiyuan
and Guo, Yongqiang
and Ou, Yuan
and Wang, Yuduan
and Gong, Yue
and Zou, Yuheng
and He, Yujia
and Xiong, Yunfan
and Luo, Yuxiang
and You, Yuxiang
and Liu, Yuxuan
and Zhou, Yuyang
and Zhu, Y. X.
and Huang, Yanping
and Li, Yaohui
and Zheng, Yi
and Zhu, Yuchen
and Ma, Yunxian
and Tang, Ying
and Zha, Yukun
and Yan, Yuting
and Ren, Z. Z.
and Ren, Zehui
and Sha, Zhangli
and Fu, Zhe
and Xu, Zhean
and Xie, Zhenda
and Zhang, Zhengyan
and Hao, Zhewen
and Ma, Zhicheng
and Yan, Zhigang
and Wu, Zhiyu
and Gu, Zihui
and Zhu, Zijia
and Liu, Zijun
and Li, Zilin
and Xie, Ziwei
and Song, Ziyang
and Pan, Zizheng
and Huang, Zhen
and Xu, Zhipeng
and Zhang, Zhongyu
and Zhang, Zhen},
title={DeepSeek-R1 incentivizes reasoning in LLMs through reinforcement learning},
journal={Nature},
year={2025},
}

@article{chu2025gpg,
  title={GPG: A Simple and Strong Reinforcement Learning Baseline for Model Reasoning},
  author={Chu, Xiangxiang and Huang, Hailang and Zhang, Xiao and Wei, Fei and Wang, Yong},
  journal={arXiv preprint arXiv:2504.02546},
  year={2025}
}

@article{schulman2017proximal,
  title={Proximal policy optimization algorithms},
  author={Schulman, John and Wolski, Filip and Dhariwal, Prafulla and Radford, Alec and Klimov, Oleg},
  journal={arXiv preprint arXiv:1707.06347},
  year={2017}
}

@article{yu2025dapo,
  title={Dapo: An open-source llm reinforcement learning system at scale},
  author={Yu, Qiying and Zhang, Zheng and Zhu, Ruofei and Yuan, Yufeng and Zuo, Xiaochen and Yue, Yu and Fan, Tiantian and Liu, Gaohong and Liu, Lingjun and Liu, Xin and others},
  journal={arXiv preprint arXiv:2503.14476},
  year={2025}
}

@article{ouyang2022training,
  title={Training language models to follow instructions with human feedback},
  author={Ouyang, Long and Wu, Jeffrey and Jiang, Xu and Almeida, Diogo and Wainwright, Carroll and Mishkin, Pamela and Zhang, Chong and Agarwal, Sandhini and Slama, Katarina and Ray, Alex and others},
  journal={Advances in Neural Information Processing Systems},
  year={2022}
}

@article{ziegler2019fine,
  title={Fine-tuning language models from human preferences},
  author={Ziegler, Daniel M and Stiennon, Nisan and Wu, Jeffrey and Brown, Tom B and Radford, Alec and Amodei, Dario and Christiano, Paul and Irving, Geoffrey},
  journal={arXiv preprint arXiv:1909.08593},
  year={2019}
}

@article{dang2025reinforcement,
  title={Reinforcement Learning for Reasoning in Small LLMs: What Works and What Doesn't},
  author={Dang, Quy-Anh and Ngo, Chris},
  journal={arXiv preprint arXiv:2503.16219},
  year={2025}
}

@article{hendrycks2021measuring,
  title={Measuring mathematical problem solving with the math dataset},
  author={Hendrycks, Dan and Burns, Collin and Kadavath, Saurav and Arora, Akul and Basart, Steven and Tang, Eric and Song, Dawn and Steinhardt, Jacob},
  journal={Advances in Neural Information Processing Systems (Datasets and Benchmarks Track)},
  year={2021}
}

@article{lewkowycz2022solving,
  title={Solving quantitative reasoning problems with language models},
  author={Lewkowycz, Aitor and Andreassen, Anders and Dohan, David and Dyer, Ethan and Michalewski, Henryk and Ramasesh, Vinay and Slone, Ambrose and Anil, Cem and Schlag, Imanol and Gutman-Solo, Theo and others},
  journal={Advances in Neural Information Processing Systems},
  year={2022}
}

@article{huang2024olympicarena,
  title={Olympicarena: Benchmarking multi-discipline cognitive reasoning for superintelligent ai},
  author={Huang, Zhen and Wang, Zengzhi and Xia, Shijie and Li, Xuefeng and Zou, Haoyang and Xu, Ruijie and Fan, Run-Ze and Ye, Lyumanshan and Chern, Ethan and Ye, Yixin and others},
  journal={Advances in Neural Information Processing Systems},
  year={2024}
}

@misc{Still-3-1.5b-preview,
    title = {Still-3-1.5b-preview: Enhancing slow thinking abilities of small models through reinforcement learning},
    author = {RUCAIBoxSTILL Team},
    year = {2025},
    howpublished = {\url{https://github.com/RUCAIBox/Slow_Thinking_with_LLMs}}
}

@article{cui2025process,
  title={Process reinforcement through implicit rewards},
  author={Cui, Ganqu and Yuan, Lifan and Wang, Zefan and Wang, Hanbin and Li, Wendi and He, Bingxiang and Fan, Yuchen and Yu, Tianyu and Xu, Qixin and Chen, Weize and others},
  journal={arXiv preprint arXiv:2502.01456},
  year={2025}
}

@inproceedings{Qwne2.5-7b-simplerl,
      title={SimpleRL-Zoo: Investigating and Taming Zero Reinforcement Learning for Open Base Models in the Wild}, 
      author={Weihao Zeng and Yuzhen Huang and Qian Liu and Wei Liu and Keqing He and Zejun Ma and Junxian He},
      booktitle={Proceedings of the Conference On Language Modeling},
      year={2025}
}

@inproceedings{kingma2014adam,
  title={Adam: A method for stochastic optimization},
  author={Kingma, Diederik P and Ba, Jimmy},
  booktitle={The Third International Conference on Learning Representations},
  year={2015}
}

@article{touvron2023llama1,
  title={Llama: Open and efficient foundation language models},
  author={Touvron, Hugo and Lavril, Thibaut and Izacard, Gautier and Martinet, Xavier and Lachaux, Marie-Anne and Lacroix, Timoth{\'e}e and Rozi{\`e}re, Baptiste and Goyal, Naman and Hambro, Eric and Azhar, Faisal and others},
  journal={arXiv preprint arXiv:2302.13971},
  year={2023}
}

@article{yang2024qwen2math,
  title={Qwen2. 5-math technical report: Toward mathematical expert model via self-improvement},
  author={Yang, An and Zhang, Beichen and Hui, Binyuan and Gao, Bofei and Yu, Bowen and Li, Chengpeng and Liu, Dayiheng and Tu, Jianhong and Zhou, Jingren and Lin, Junyang and others},
  journal={arXiv preprint arXiv:2409.12122},
  year={2024}
}

@article{roelofs2019meta,
  title={A meta-analysis of overfitting in machine learning},
  author={Roelofs, Rebecca and Shankar, Vaishaal and Recht, Benjamin and Fridovich-Keil, Sara and Hardt, Moritz and Miller, John and Schmidt, Ludwig},
  journal={Advances in Neural Information Processing Systems},
  year={2019}
}

@inproceedings{yu2022understanding,
  title={Understanding robust overfitting of adversarial training and beyond},
  author={Yu, Chaojian and Han, Bo and Shen, Li and Yu, Jun and Gong, Chen and Gong, Mingming and Liu, Tongliang},
  booktitle={International Conference on Machine Learning},
  year={2022},
}

@article{everitt2021reward,
  title={Reward tampering problems and solutions in reinforcement learning: A causal influence diagram perspective},
  author={Everitt, Tom and Hutter, Marcus and Kumar, Ramana and Krakovna, Victoria},
  journal={Synthese},
  year={2021},
  publisher={Springer}
}

@article{pan2022effects,
  title={The effects of reward misspecification: Mapping and mitigating misaligned models},
  author={Pan, Alexander and Bhatia, Kush and Steinhardt, Jacob},
  journal={arXiv preprint arXiv:2201.03544},
  year={2022}
}

@article{liu2025understanding,
  title={Understanding r1-zero-like training: A critical perspective},
  author={Liu, Zichen and Chen, Changyu and Li, Wenjun and Qi, Penghui and Pang, Tianyu and Du, Chao and Lee, Wee Sun and Lin, Min},
  journal={arXiv preprint arXiv:2503.20783},
  year={2025}
}

@article{hu2025open,
  title={Open-reasoner-zero: An open source approach to scaling up reinforcement learning on the base model},
  author={Hu, Jingcheng and Zhang, Yinmin and Han, Qi and Jiang, Daxin and Zhang, Xiangyu and Shum, Heung-Yeung},
  journal={arXiv preprint arXiv:2503.24290},
  year={2025}
}

@misc{lighteval,
  author = {Habib, Nathan and Fourrier, Clémentine and Kydlíček, Hynek and Wolf, Thomas and Tunstall, Lewis},
  title = {LightEval: A lightweight framework for LLM evaluation},
  year = {2023},
  version = {0.8.0},
  url = {https://github.com/huggingface/lighteval}
}

@article{chen2025empirical,
  title={An Empirical Study on Eliciting and Improving R1-like Reasoning Models},
  author={Chen, Zhipeng and Min, Yingqian and Zhang, Beichen and Chen, Jie and Jiang, Jinhao and Cheng, Daixuan and Zhao, Wayne Xin and Liu, Zheng and Miao, Xu and Lu, Yang and others},
  journal={arXiv preprint arXiv:2503.04548},
  year={2025}
}

@article{zhao2023survey,
  title={A survey of large language models},
  author={Zhao, Wayne Xin and Zhou, Kun and Li, Junyi and Tang, Tianyi and Wang, Xiaolei and Hou, Yupeng and Min, Yingqian and Zhang, Beichen and Zhang, Junjie and Dong, Zican and others},
  journal={arXiv preprint arXiv:2303.18223},
  year={2023}
}

@misc{qwen2024qwq32b,
  title        = {{QwQ: Reflect Deeply on the Boundaries of the Unknown}},
  author       = {{Qwen Team}},
  year         = {2024},
  howpublished = {\url{https://qwenlm.github.io/blog/qwq-32b-preview/}},
}

@misc{openai2024o1,
  title        = {{Introducing OpenAI O1 Preview}},
  author       = {{OpenAI}},
  year         = {2024},
  howpublished = {\url{https://openai.com/index/introducing-openai-o1-preview/}},
}

@inproceedings{schulman2015high,
  title={High-dimensional continuous control using generalized advantage estimation},
  author={Schulman, John and Moritz, Philipp and Levine, Sergey and Jordan, Michael and Abbeel, Pieter},
  booktitle={The Fourth International Conference on Learning Representations},
  year={2016}
}

@article{williams1992simple,
  title={Simple statistical gradient-following algorithms for connectionist reinforcement learning},
  author={Williams, Ronald J},
  journal={Machine learning},
  year={1992},
  publisher={Springer}
}

@article{sutton1999policy,
  title={Policy gradient methods for reinforcement learning with function approximation},
  author={Sutton, Richard S and McAllester, David and Singh, Satinder and Mansour, Yishay},
  journal={Advances in Neural Information Processing Systems},
  year={1999}
}

@inproceedings{schulman2015trust,
  title={Trust region policy optimization},
  author={Schulman, John and Levine, Sergey and Abbeel, Pieter and Jordan, Michael and Moritz, Philipp},
  booktitle={International Conference on Machine Learning},
  year={2015},
}

@inproceedings{haarnoja2018soft,
  title={Soft actor-critic: Off-policy maximum entropy deep reinforcement learning with a stochastic actor},
  author={Haarnoja, Tuomas and Zhou, Aurick and Abbeel, Pieter and Levine, Sergey},
  booktitle={International Conference on Machine Learning},
  year={2018},
}

@article{shannon1948mathematical,
  title={A mathematical theory of communication},
  author={Shannon, Claude E},
  journal={The Bell system technical journal},
  year={1948},
  publisher={Nokia Bell Labs}
}

@incollection{robbins1971convergence,
  title={A convergence theorem for non negative almost supermartingales and some applications},
  author={Robbins, Herbert and Siegmund, David},
  booktitle={Optimizing methods in statistics},
  year={1971},
  publisher={Elsevier}
}

@inproceedings{kwon2023efficient,
  title={Efficient memory management for large language model serving with pagedattention},
  author={Kwon, Woosuk and Li, Zhuohan and Zhuang, Siyuan and Sheng, Ying and Zheng, Lianmin and Yu, Cody Hao and Gonzalez, Joseph and Zhang, Hao and Stoica, Ion},
  booktitle={Proceedings of the 29th symposium on operating systems principles},
  year={2023}
}

@article{yue2025does,
  title={Does reinforcement learning really incentivize reasoning capacity in llms beyond the base model?},
  author={Yue, Yang and Chen, Zhiqi and Lu, Rui and Zhao, Andrew and Wang, Zhaokai and Song, Shiji and Huang, Gao},
  journal={Advances in Neural Information Processing Systems},
  year={2025}
}

@inproceedings{hendrycks2020measuring,
  title={Measuring massive multitask language understanding},
  author={Hendrycks, Dan and Burns, Collin and Basart, Steven and Zou, Andy and Mazeika, Mantas and Song, Dawn and Steinhardt, Jacob},
  booktitle={The Twelfth International Conference on Learning Representations},
  year={2021}
}

@article{jain2024livecodebench,
  title={Livecodebench: Holistic and contamination free evaluation of large language models for code},
  author={Jain, Naman and Han, King and Gu, Alex and Li, Wen-Ding and Yan, Fanjia and Zhang, Tianjun and Wang, Sida and Solar-Lezama, Armando and Sen, Koushik and Stoica, Ion},
  journal={arXiv preprint arXiv:2403.07974},
  year={2024}
}

\appendix
\section{Algorithm}
\textbf{\sysname Training}\quad The training process of \sysname can be summarized as follows: 
1) Sample two groups of responses, $O_\theta$ and $O_{r e f}$, of equal size from both the policy model $\pi_\theta$ and the reference model $\pi_{r e f}$, respectively. 
2) For each response in $O_\theta$, compute its group relative advantage and the advantage margin with the corresponding response in $O_{r e f}$. 
The values, calculated by~\eqref{eq:5}, are then used to perform gradient updates according to~\eqref{eq:6}.
It is important to note that during the training process, the parameters of the reference model $\pi_{r e f}$ remain frozen and do not undergo gradient updates.
The training procedure is described in Algorithm \ref{alg:aapo}.
\begin{algorithm*}[t]
\caption{Advantage-Augmented Policy Optimization}
\label{alg:aapo}

\KwIn{policy model $\pi_\theta$, reference model $\pi_{r e f}$, group size $G$, reward functions $\scriptstyle F=\{\text{format}, \text{accuracy},\cdots\}$, reward functions' corresponding weights $\scriptstyle W=\{w_{\text{format}}, w_{\text{accuracy}}, \cdots\}$, data batch $\mathcal{B}$, total training steps $\mathcal{S}_{gloabl}$.}

\BlankLine
\ForEach{global training step $\mathcal{S} < \mathcal{S}_{gloabl}$}{

    \ForEach{data in data batch $\mathcal{B}$}{
        Sample $O_{\theta}=\{o_{\theta_1}, o_{\theta_2}, \cdots, o_{\theta_G}\}$ from $\pi_\theta$;
        
        Sample $O_{r e f}=\{o_{r e f_{1}}, o_{r e f_{2}}, \cdots, o_{r e f_{G}}\}$ from $\pi_{r e f}$;
        
        Compute rewards $\scriptstyle R_\theta=\{R_{\theta}^{\text{format}}, R_{\theta}^{\text{accuracy}}, \cdots, R_{\theta}^{\cdots}\}$ and $\scriptstyle R_{r e f}=\{R_{r e f}^{\text{format}}, R_{r e f}^{\text{accuracy}}, \cdots, R_{r e f}^{\cdots}\}$ for each response in $O_{\theta}$ and $O_{r e f}$ using the reward functions in $F$;

        Compute the weighted rewards $\scriptstyle R_\theta^{\text{weighted}}=R_\theta W^{\top}$ and $\scriptstyle R_{ref}^{\text{weighted}}=R_{r e f} W^{\top}$;

        Compute the augmented advantage $\hat{A}_{i, t}^*$ following Equation~\eqref{eq:5} for each sample in $O_{\theta}$;

        Compute loss for $O_{\theta}$ following Equation~\eqref{eq:4};
    }
    Update $\pi_\theta$ following Equation~\eqref{eq:6};
}
\end{algorithm*}
\section{Proof}
\label{sec:proof}
\textbf{Assumption} \quad \textit{Here, we state all assumptions we use in our proof of theorem.
(1) The gradient of the log-policy is bounded: $||\nabla_{\theta}\log\pi_{\theta}(o)||\leq M$, and $-\log\pi_{\theta}(o)$ is $L_0$-Lipschitz.
(2) Assume that the stochastic gradient is unbiased and that the per-sample gradient has bounded second moment.
(3) For all $\theta$, the per-sample loss $\ell(q,o;\theta)=-\log \pi_\theta(o\mid q)\,\hat A^*$ has a gradient with bounded second moment:
$
\sup_{\theta} \mathbb{E}_{(q,o)\sim\pi_\theta}\!\left[\left\|\nabla_\theta \ell(q,o;\theta)\right\|^{2}\right] \le \sigma^2
$, where $\sigma^2$ is a constant.
}

\textbf{Theorem 1.} \textit{(Stability) \quad Since the rewards are bounded, the group standard deviation satisfies $0\leq\sigma_{min}\leq\sigma$, and the log-likelihood gradients are bounded as $||\nabla_{\theta}\log\pi_{\theta}(o)||\leq M$.
Then, each gradient step with learning rate $\eta_k$ satisfies $||\theta_{k+1}-\theta_k||\leq\eta_{k}MB$, where $B=\frac{R_{max}-R_{min}}{\sigma_{min}}+\operatorname{max}(|\delta_\mathrm{low}|, |\delta_\mathrm{high}|)$ is the uniform bound on the \sysname weights.
The expected objective is bounded from $\mathcal{L}(\theta)\geq-B\log|\mathcal{V}|$, where $|\mathcal{V}|$ is the vocabulary size.
Hence, \sysname training is stable: the objective cannot diverge to $-\infty$ and parameter updates are always finite.}
\begin{proof}
We restate the definitions used: 
For a group $\mathcal{G}$, the gradient of the empirical loss is: 
$$
\nabla_{\theta}\mathcal{L}_{\mathcal{G}}(\theta)=\frac{1}{N_{\mathcal{G}}}\sum_{o\in O}\nabla_{\theta}[-\log\pi_{\theta}(o)]\hat{A}^* .
$$
By assumption (1), we have
$$
||\nabla_{\theta}\mathcal{L}_{\mathcal{G}}(\theta)||\leq\frac{1}{N_{\mathcal{G}}}\sum_{o \in O}M\cdot B=MB .
$$
Thus, one gradient update with learning rate $\eta_k$ yields
$$
    ||\theta_{k+1}-\theta_k||=\eta_{k}||\nabla_{\theta}\mathcal{L}_{\mathcal{G}}(\theta)||\leq\eta_{k}MB .
$$
For any response $o$, 
$$
\mathbb{E}_{a \sim \pi_\theta} \left[ - \log \pi_\theta(o) \right] = H(\pi_\theta(\cdot \mid o_{<t})) \leq \log|\mathcal{V}| .
$$
where $H(\cdot)=-\sum_{x}p(x)\log p(x)$ denotes Shannon entropy \citep{shannon1948mathematical}. 
Multiplying by the bounded advantage $|\hat{A}^*|\leq B$ and averaging over tokens in the batch, we obtain
$$
    \mathcal{L}(\theta)\geq-B\log|\mathcal{V}| .
$$
we conclude that parameter updates are bounded and the objective is lower bounded (which means the objective cannot diverge to $-\infty$). 
Therefore, \sysname training is stable and cannot diverge.
\end{proof}

\textbf{Theorem 2.} \textit{(Convergence) \quad By assumption (2), let the step sizes satisfy the Robbins–Monro conditions $\eta_k>0$, $\sum_{k}\eta_k=\infty$, $\sum_{k}\eta_k^2<\infty$. 
\sysname converges to a stationary point of its expected objective $\liminf_{k \to \infty} \mathbb{E}\left[ \left\| \nabla \mathcal{L}(\theta_k) \right\|^2 \right] = 0$.
Moreover, if a constant step size $\eta<\frac{1}{BL_0}$ is used, where $L_0$ is the smoothness constant of $-\log\pi_{\theta}(o)$, then the iterates converge to a neighborhood of stationarity $\limsup_{K \to \infty} \frac{1}{K}\sum_{k=1}^K \mathbb{E}\!\left[ \left\| \nabla \mathcal{L}(\theta_k) \right\|^2 \right] \;\lesssim\; \mathcal{O}(\eta) + \mathcal{O}\!\left(\tfrac{1}{N_\mathcal{G}}\right)$.}
\begin{proof}
We adopt the same definitions as in the Proof of Theorem 1.
Since $-\log\pi_{\theta}(o\mid q)$ is $L_0$-smooth (i.e., its gradient is $L_0$-Lipschitz) and the advantage is bounded by $B$, the empirical loss $\mathcal{L}_{\mathcal{G}}(\theta)$ is $L= B L_0$-smooth:
\[
    \left\| \nabla \mathcal{L}_{\mathcal{G}}(\theta) - \nabla \mathcal{L}_{\mathcal{G}}(\theta') \right\|_2 \le L \left\| \theta - \theta' \right\|_2 .
\]
By the descent lemma for $L$-smooth functions, for any step size $\eta_k$,
\[
    \mathcal{L}_{\mathcal{G}}(\theta - \eta_k g) \le \mathcal{L}_{\mathcal{G}}(\theta) - \eta_k \langle g, \nabla \mathcal{L}_{\mathcal{G}}(\theta) \rangle + \tfrac{L}{2}\eta_k^2 \|g\|^2 .
\]
Choosing $g=\nabla \mathcal{L}_{\mathcal{G}}(\theta_k)$ and requiring $\eta_k\le 1/L$, we obtain
\[
    \mathcal{L}_{\mathcal{G}}(\theta_{k+1}) \le \mathcal{L}_{\mathcal{G}}(\theta_k) - \tfrac{\eta_k}{2} \left\| \nabla \mathcal{L}_{\mathcal{G}}(\theta_k) \right\|^2 .
\]

By assumption (3), for all $\theta$, the per-sample loss $\ell(q,o;\theta)=-\log \pi_\theta(o\mid q)\,\hat A^*$ has a gradient with bounded second moment:
\[
\sup_{\theta} \mathbb{E}_{(q,o)\sim\pi_\theta}\!\left[\left\|\nabla_\theta \ell(q,o;\theta)\right\|^{2}\right] \le \sigma^2 ,
\]
where $\sigma^2$ is a constant. Since $\nabla \mathcal{L}_{\mathcal{G}}(\theta_k)$ is the average of $N_{\mathcal G}$ i.i.d.\ samples conditional on $\theta_k$, we have
\begin{gather}\notag
\mathbb{E}\!\left[\nabla \mathcal{L}_{\mathcal{G}}(\theta_k)\right] = \nabla \mathcal{L}(\theta_k),\notag\\
\mathbb{E}\!\left[\left\|\nabla \mathcal{L}_{\mathcal{G}}(\theta_k)-\nabla \mathcal{L}(\theta_k)\right\|^{2}\right] \le \frac{\sigma^2}{N_{\mathcal G}}.\notag  
\end{gather}
Consequently (by variance decomposition),
\[
\mathbb{E}\!\left[\left\|\nabla \mathcal{L}_{\mathcal{G}}(\theta_k)\right\|^{2}\right]
\le \left\|\nabla \mathcal{L}(\theta_k)\right\|^{2} + \frac{\sigma^2}{N_{\mathcal G}} .
\]

Taking total expectation and noting $\mathbb{E}[\mathcal{L}_{\mathcal{G}}(\theta)]=\mathcal{L}(\theta)$, we obtain
\begin{equation}\notag
\scalebox{0.9}{$\displaystyle
\begin{aligned}
\mathbb{E}\!\left[\mathcal{L}(\theta_{k+1})\right]
\le \mathbb{E}\!\left[\mathcal{L}(\theta_k)\right] - \tfrac{\eta_k}{2}\,\mathbb{E}\!\left[\left\|\nabla \mathcal{L}(\theta_k)\right\|^{2}\right]
+ \tfrac{L}{2}\,\tfrac{\eta_k^{2}\sigma^2}{N_{\mathcal G}}.
\end{aligned}
$}
\end{equation}
Since $\mathcal{L}(\theta)$ is lower bounded (Theorem 1), applying the Robbins--Siegmund theorem ~\citep{robbins1971convergence} gives
\[
\sum_{k=0}^{\infty} \eta_k \,\mathbb{E}\!\left[\left\| \nabla \mathcal{L}(\theta_k) \right\|^{2}\right] < \infty .
\]
Given $\sum_k \eta_k = \infty$, it follows that
\[
\liminf_{k\to\infty} \mathbb{E}\!\left[\left\|\nabla \mathcal{L}(\theta_k)\right\|^{2}\right] = 0 .
\]

If $\eta_k\equiv \eta \le 1/L$ is fixed, the residual is of order $\mathcal{O}(\eta)$ (from smoothness) and $\mathcal{O}(1/N_{\mathcal G})$ (from gradient variance). 
Hence
\begin{equation}\notag
\scalebox{0.94}{$\displaystyle
\begin{aligned}
\limsup_{K\to\infty} \frac{1}{K}\sum_{k=1}^{K} \mathbb{E}\!\left[\left\|\nabla \mathcal{L}(\theta_k)\right\|^{2}\right]
\;\lesssim\; \mathcal{O}(\eta) + \mathcal{O}\!\left(\tfrac{1}{N_{\mathcal G}}\right).
\end{aligned}
$}
\end{equation}
With diminishing step sizes, the algorithm converges to a stationary point of the expected objective; with a small constant step size, it converges to a neighborhood of stationarity whose size depends on $\eta$ and the group size. The advantage margin does not affect the asymptotic rates, as it only changes the constant $B$ (and hence $L=BL_0$).
\end{proof}

\section{Experiment details}
\label{sec: train}
\subsection{Reward rules}
\textbf{Format Reward} \quad To encourage adherence to structured reasoning, we adopt a binary format reward $R_{format}(o) \in \{0, 1\}$, which assigns a reward of 1 if the model response 
$o$ conforms to the expected template by containing the delimiter sequence ``\textbackslash n \textless /think\textgreater \textbackslash n", and 0 otherwise.\\
\textbf{Cosine Scaled Reward} \quad We adopt the $R_{cosine\_scaled\_accuracy} \in \{0, 1\}$ as expressed in equation \eqref{eq:9}, which encourages correct outputs with shorter lengths and penalizes incorrect outputs with reduced severity as their length increases, following a cosine annealing schedule in equation \eqref{eq:10}.
\begin{equation}\label{eq:9}
\scalebox{0.8}{$\displaystyle
R_{cosine\_scaled\_accuracy}(o) = \begin{cases}
R_{\text{correct}}(l), & \text{if correct} \\
R_{\text{wrong}}(l), & \text{if wrong}
\end{cases},
$}
\end{equation}
where
\begin{equation}\label{eq:10}
\scalebox{0.72}{$\displaystyle
\begin{aligned}
R_{\text{correct}}(l) &= \alpha_{\min}^c + \frac{1}{2}(\alpha_{\max}^c - \alpha_{\min}^c)\left[1 + \cos\left(\pi \frac{l}{L}\right)\right], \\
R_{\text{wrong}}(l) &= \alpha_{\max}^w + \frac{1}{2}(\alpha_{\min}^w - \alpha_{\max}^w)\left[1 + \cos\left(\pi \frac{l}{L}\right)\right],
\end{aligned}.
$}
\end{equation}\\
\textbf{Accuracy Reward} \quad We adopt a standard accuracy reward $R_{accuracy} \in \{0, 1\}$, which assigns a binary reward of 1 for correct response and 0 otherwise, providing a sparse but direct reward signal.
\subsection{Training setup}
\textbf{Training DeepSeek-R1-Distill-Qwen-1.5B} \quad For training DeepSeek-R1-Distill-Qwen-1.5B on open-rs experiment \citep{dang2025reinforcement}, we adopt the $R_{format}$ and $R_{cosine\_scaled\_accuracy}$ reward functions, with respective weights of 1 and 2.
We directly perform our \sysname training from the base model without any SFT with the group size of 6 and the per\_device\_batch\_size of 6 with gradient\_accumulation\_steps of 4 on 2 Nvidia A800 GPUs with 80G VRAM.
The system prompt adopted in the RL training is provided below.
\begin{tcolorbox}[
    title=Training prompt \sysname-1.5B model,
    coltitle=black,
    colback=white, 
    colframe=blue!20, 
    boxrule=0.4mm, 
    arc=1mm 
]
A conversation between User and Assistant.
The user asks a question, and the Assistant solves it.
The assistant first thinks about the reasoning process in the mind and then provides the user with the answer, and put your final answer within \textbackslash\textbackslash boxed{{}} . 
The reasoning process and answer are enclosed within \textless think\textgreater \textless/think \textgreater and \textless answer\textgreater \textless/answer\textgreater tags, respectively, i.e., \textless think\textgreater reasoning process here \textless/think \textgreater \textless answer\textgreater answer here \textless/answer\textgreater.
Note that respond by English, NOT use other languages.
\end{tcolorbox}
\textbf{Training Qwen2.5-Math-7B} \quad For training Qwen2.5-Math-7B \citep{yang2024qwen2math} on  simplelr\_qwen\_level3to5 \citep{Qwne2.5-7b-simplerl}, we adopt the $R_{accuracy}$ reward function with a weight of 1.
We directly perform our \sysname training without any SFT with the group size of 8 and the per\_device\_batch\_size 8 with gradient\_accumulation\_steps of 4 on 8 Nvidia A800 GPUs with 80G VRAM.
The system prompt adopted in the RL training is provided below.
\begin{tcolorbox}[
    title=Training prompt for \sysname-7B model,
    coltitle=black,
    colback=white,
    colframe=blue!20,
    boxrule=0.4mm,
    arc=1mm
]
You are a helpful AI Assistant, designed to provided well-reasoned anddetailed responses.
You FIRST think about the reasoning process as an internal monologue and then provide the user with the answer. 
The reasoning process MUST BE enclosed within \textless think\textgreater and \textless /think\textgreater tags.
\end{tcolorbox}
\textbf{Training Llama series models} \quad For training Llama-3.2-1B-Instruct and Llama-3.2-3B-Instruct models, we adopt the simplelr\_able\_level3to5 dataset \citep{Qwne2.5-7b-simplerl} and the $R_{accuracy}$ reward function with a weight of 1.
We directly perform our \sysname training without any SFT with the group size of 8 and the per\_device\_batch\_size of 16 on 4 Nvidia A800 GPUs with 80G VRAM.
There is no extra system prompt added when training Llama series models.
\subsection{Evaluation setup}
\label{sec: eval}
\textbf{Evaluation of Qwen series models} \quad All the reproduced results in Table \ref{tab:1} are evaluated using the lighteval framework \citep{lighteval} with vllm backend proposed by Hugging Face.
Our evaluation experiments on all benchmarks are conducted on a single Nvidia A800 GPU with 80G VRAM, with system prompt provided in below.
\begin{tcolorbox}[
    title=System prompt for Qwen series evaluation on all benchmarks,
    coltitle=black,
    colback=white,
    colframe=blue!20,
    boxrule=0.4mm,
    arc=1mm
]
Solve the following math problem efficiently and clearly. 
The last line of your response should be of the following format: `Therefore, the final answer is: \$\textbackslash\textbackslash boxed\{\{ ANSWER\}\}\$. 
I hope it is correct' (without quotes) where ANSWER is just the final number or expression that solves the problem. 
Think step by step before answering.
\end{tcolorbox}

\textbf{Evaluation of Llama series models} \quad All the reported results in Table \ref{tab:1} are evaluated using vllm backend for generation and the evaluation script provided by \citet{Qwne2.5-7b-simplerl}.
Our evaluation experiments on all benchmarks are conducted on 4 Nvidia A800 GPUs with 80G VRAM.
\begin{figure}[t]
    \centering
    \includegraphics[width=0.6\linewidth]{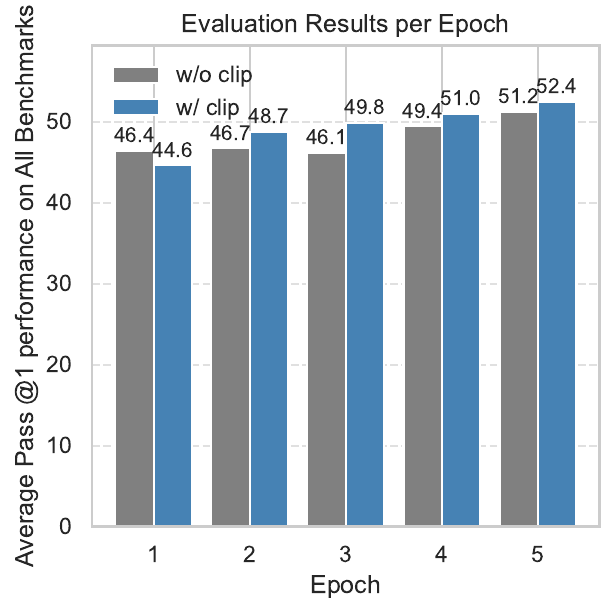}
    \caption{Ablation study on $\operatorname{clip}$ operation, reported by average pass@1 scores across five benchmarks.}
    \label{fig:clip}
\end{figure}
\section{More experiment results}
\label{appendix: more}
\subsection{Performance on Out-of-Domain benchmarks}
\label{appendix: ood}
Beyond mathematical reasoning tasks, we evaluated \sysname-1.5B and \sysname-7B on a more general benchmark: MMLU~\citep{hendrycks2020measuring}, MMLU encompasses 57 subdomains spanning STEM, social science, and other domains.
As reported in Table~\ref{tab:general}, results show that even when our method trained models using only a small subset of math problems, \sysname-1.5B achieved a slight improvement (\textbf{+0.27}) over DeepSeek-R1-Distill-Qwen-1.5B model, attributing to the base model's inherent strength.
However, \sysname-7B demonstrated substantial gains (\textbf{+13.69}) over Qwen2.5-Math-7B model.
Except for MMLU benchmark, we evaluate \sysname-1.5B on LiveCodeBench~\citep{jain2024livecodebench}, which evaluates models on a variety of code-related scenarios, such as code generation, self-repair, test output prediction, and code execution.
Since the max\_model\_length of \sysname-7B is 4,096, it is not appropriate to evaluate \sysname-7B on LiveCodeBench.
As shown in Table~\ref{tab:code}, \sysname-1.5B also outperforms its base model by \textbf{1.23} on code-related benchmark.
These findings demonstrate that our \sysname does not compromise model's generalization. 
Instead, it extends model's reasoning capabilities to other domains to a certain extent.
\begin{table}[hb]
  \centering
  \setlength{\tabcolsep}{2pt}
  \begin{adjustbox}{width=0.48\textwidth, center} 
  \begin{tabular}{lccc}
    \toprule
    LiveCodeBench&DeepSeek-R1-Distill-Qwen-1.5B&\sysname-1.5B&$\Delta$ \\
    \midrule
    Average & 25.78 & 27.01 & +1.23\\
    \bottomrule
  \end{tabular}
  \end{adjustbox}
  \caption{Performance on LiveCodeBench benchmark.}
\label{tab:code}
\end{table}
\begin{figure*}[t]
    \centering
    \begin{subfigure}[b]{0.23\textwidth}
        \centering
        \includegraphics[width=\textwidth]{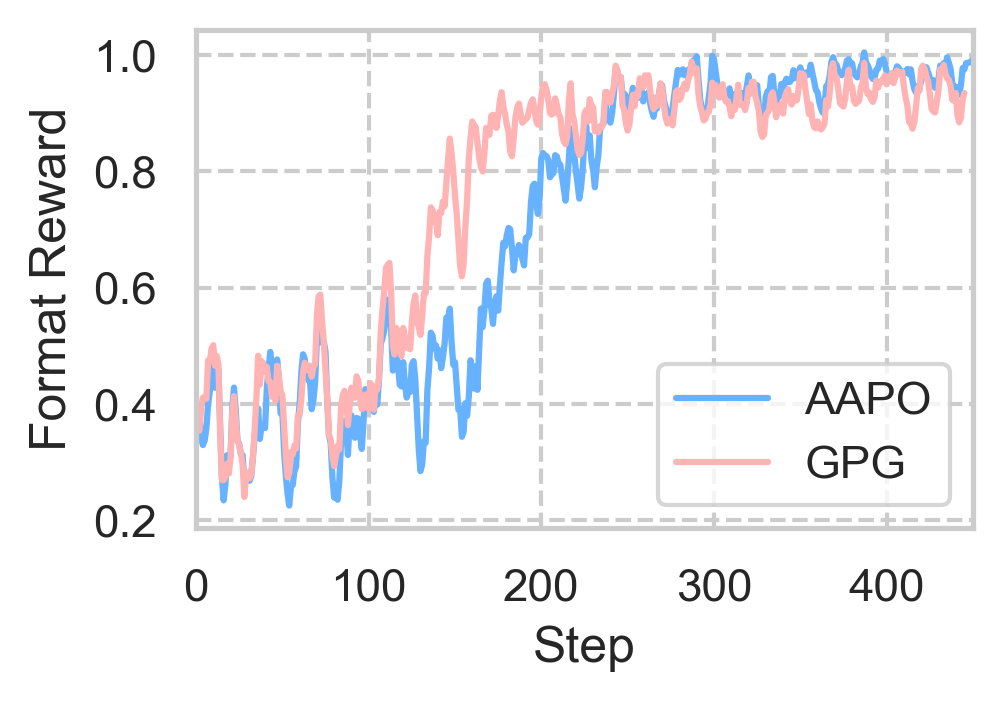}
        \label{fig:subfig1}
    \end{subfigure}%
    \hspace{0\textwidth}
    \begin{subfigure}[b]{0.23\textwidth}
        \centering
        \includegraphics[width=\textwidth]{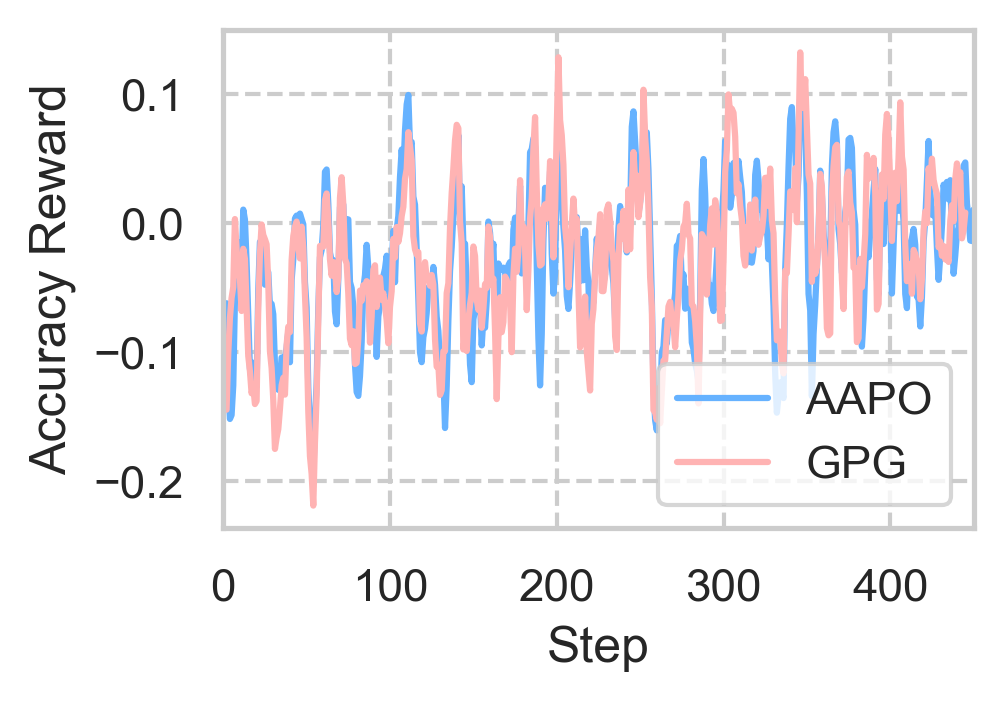}
        \label{fig:subfig2}
    \end{subfigure}%
    \hspace{0\textwidth}
    \begin{subfigure}[b]{0.23\textwidth}
        \centering
        \includegraphics[width=\textwidth]{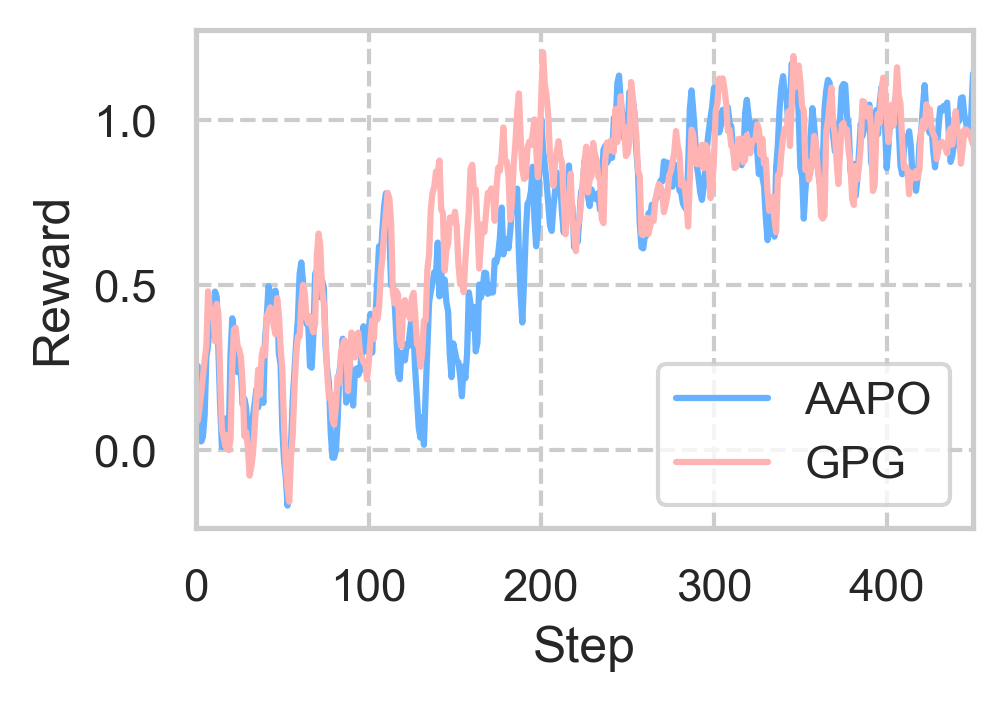}
        \label{fig:subfig3}
    \end{subfigure}
    \begin{subfigure}[b]{0.23\textwidth}
        \centering
        \includegraphics[width=\textwidth]{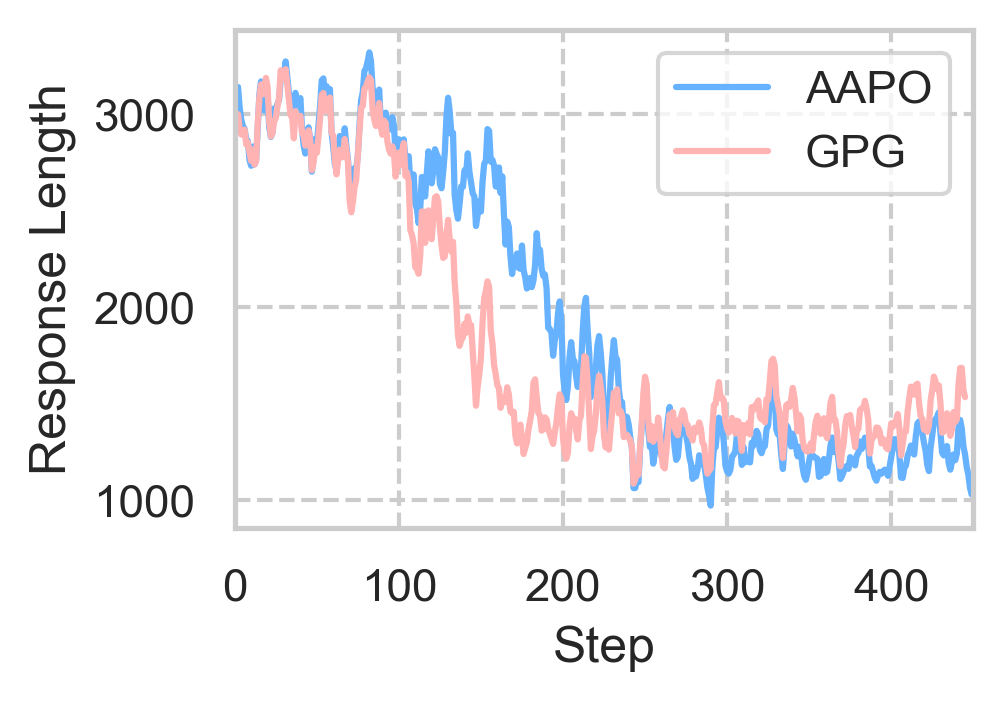}
        \label{fig:subfig4}
    \end{subfigure} \\[0em]
    \begin{subfigure}[b]{0.23\textwidth}
        \centering
        \includegraphics[width=\textwidth]{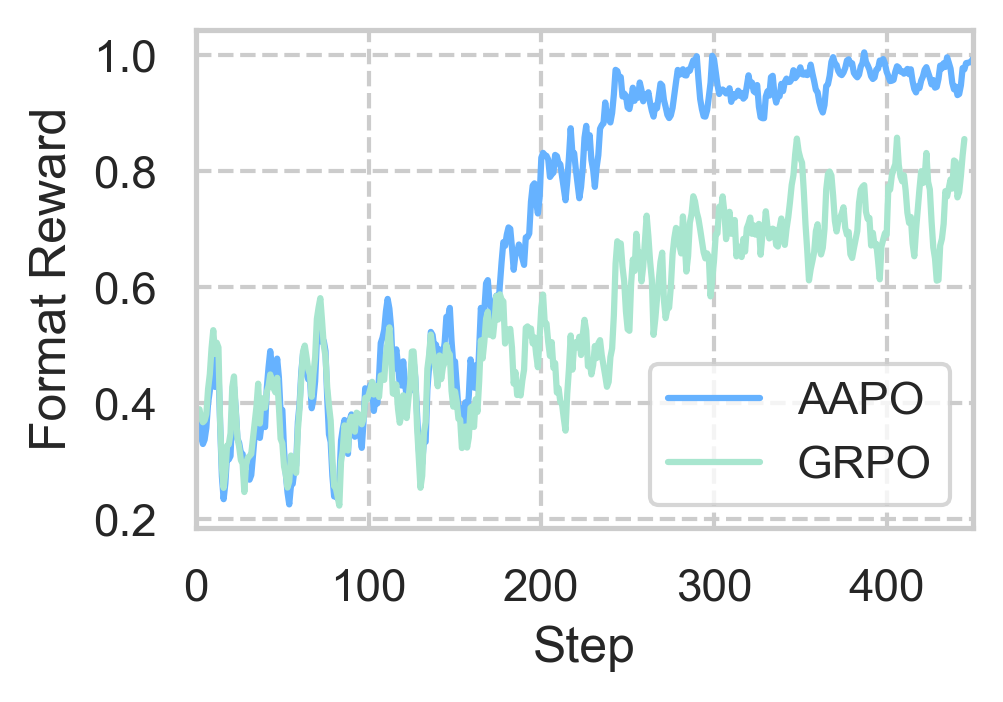}
        \label{fig:subfig5}
    \end{subfigure}%
    \hspace{0\textwidth}
    \begin{subfigure}[b]{0.23\textwidth}
        \centering
        \includegraphics[width=\textwidth]{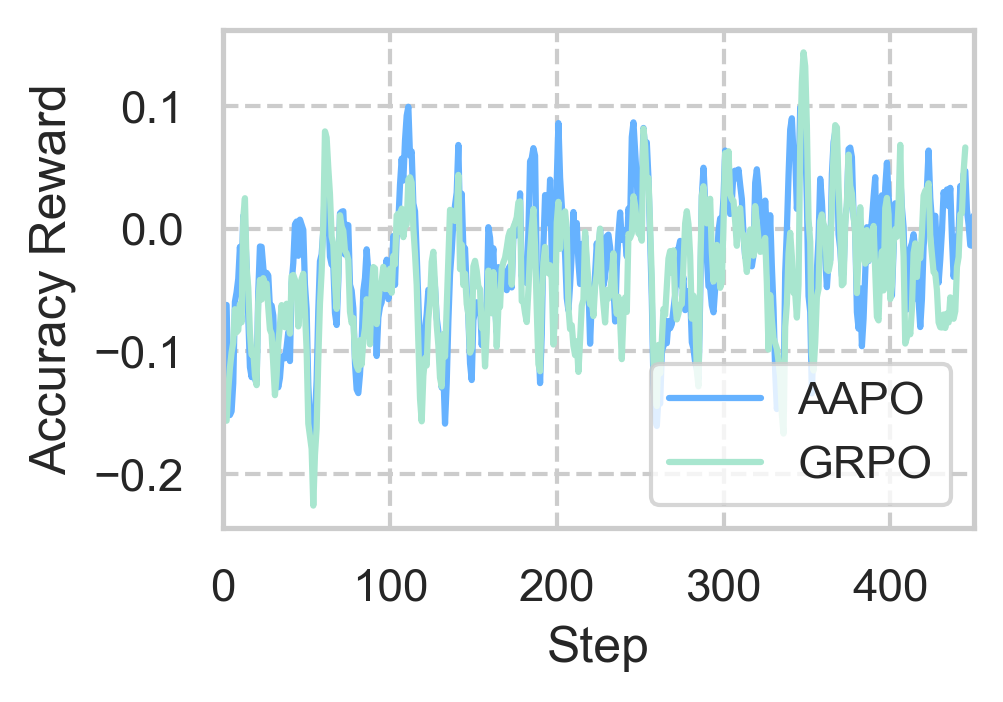}
        \label{fig:subfig6}
    \end{subfigure}%
    \hspace{0\textwidth}
    \begin{subfigure}[b]{0.23\textwidth}
        \centering
        \includegraphics[width=\textwidth]{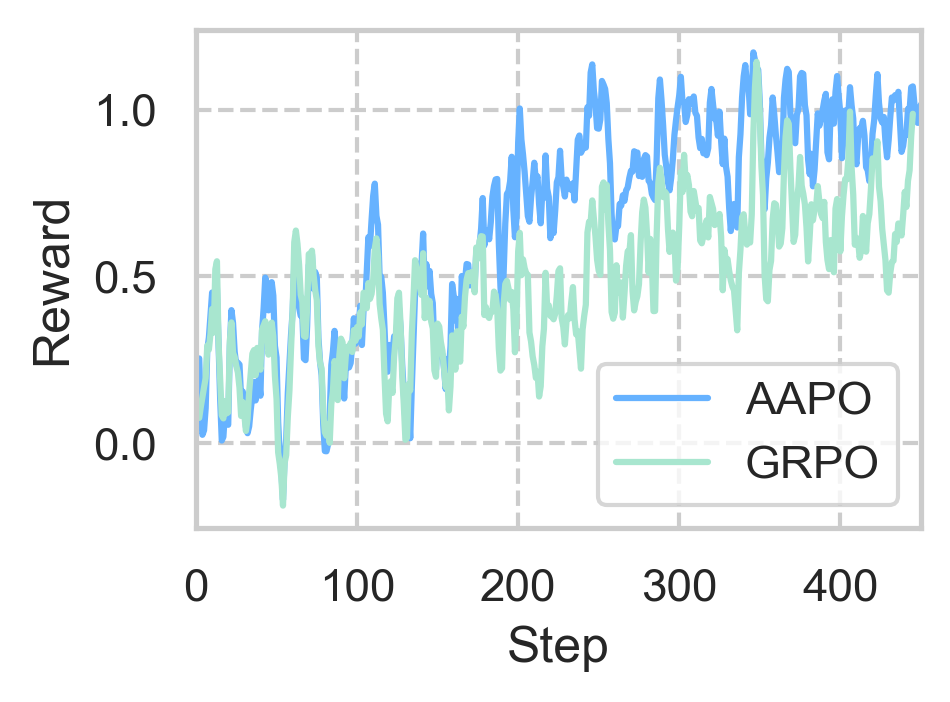}
        \label{fig:subfig7}
    \end{subfigure}
    \begin{subfigure}[b]{0.23\textwidth}
        \centering
        \includegraphics[width=\textwidth]{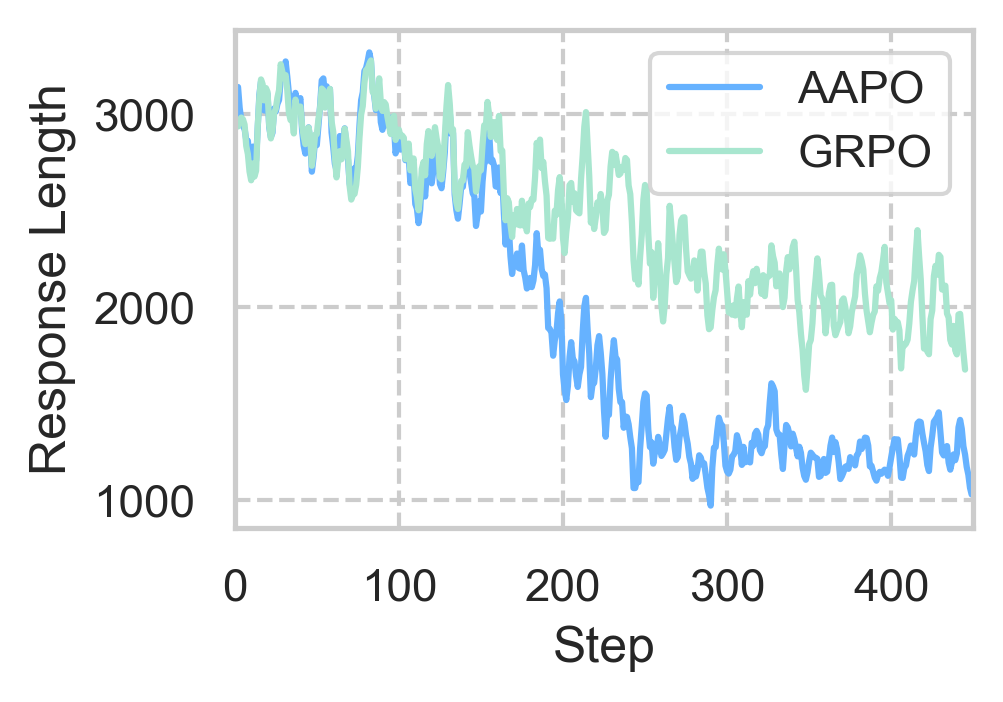}
        \label{fig:subfig8}
    \end{subfigure}
    \caption{Training process with DeepSeek-R1-Distill-Qwen-1.5B on open-rs \citep{dang2025reinforcement} utilizing our proposed \sysname algorithm.
    Compared to GPG \citep{chu2025gpg} and GRPO \citep{shao2024deepseekmath}, \sysname demonstrates better stability during training and achieves superior performance in the final results as shown in Table \ref{tab:1}.}
    \label{fig:2}
\end{figure*}
\subsection{The effect of training group size}
\label{appendix: groupsize}
To examine the effect of group size $G$ on training dynamics, we conducted a set of additional experiments with varying group sizes on DeepSeek-R1-Distill-Qwen-1.5B and Qwen2.5-Math-7B models.
The results are summarized in Table~\ref{tab:3}.
The experimental results indicate that \sysname is robust to group size across model size.
Based on this observation, we use group size $G=6$ as our main results in Table~\ref{tab:1}, which is consistent to other baselines.
\begin{table}[hb]
  \centering
  \setlength{\tabcolsep}{2pt}
  \begin{adjustbox}{width=0.48\textwidth, center} 
  \begin{tabular}{lcccccc>{\columncolor{mygray}}c}
    \toprule
    Model&G&AIME24&MATH-500&AMC23&Minerva&OlympiadBench&Avg. \\
    \midrule
    \multirow{3}{*}{\small \sysname-1.5B}
    & 2 & 36.7 & 83.0 & 80.0 & 32.0 & 51.4 & 56.6 \\
    & 6 & 33.3 & 86.0 & 80.0 & 30.9 & 53.3 & 56.7 \\
    & 10 & 36.7 & 85.6 & 77.5 & 27.6 & 52.0 & 55.9 \\
    \midrule
    \multirow{3}{*}{\small \sysname-7B}
    & 2 & 33.3 & 78.8 & 62.5 & 32.4 & 39.4 & 49.3 \\
    & 6 & 30.0 & 82.4 & 70.0 & 35.3 & 44.3 & 52.4 \\
    & 10 & 36.7 & 81.8 & 65.0 & 35.3 & 44.6 & 52.7 \\
    \bottomrule
  \end{tabular}
  \end{adjustbox}
  \caption{Additional ablation study on group size $G$.}
\label{tab:3}
\end{table}
\subsection{Pass@k performance}
\label{appendix: passk}
\begin{table}[b]
  \centering
  \setlength{\tabcolsep}{2pt}
  \begin{adjustbox}{width=0.48\textwidth, center} 
  \begin{tabular}{lcccccc}
    \toprule
    Model&K&AIME24&MATH-500&AMC23&Minerva&OlympiadBench \\
    \midrule
    \multirow{4}{*}{\small DeepSeek-R1-Distill-Qwen-1.5B}
     & 8 & 64.3 & 95.6 & 93.5 & 48.2 & 72.2 \\
     & 16 & 73.2 & 97.1 & 95.4 & 52.1 & 76.4 \\
     & 32 & 78.9 & 98.1 & 97.4 & 55.2 & 79.7 \\
     & 64 & 83.3 & 99.0 & 100.0 & 57.1 & 81.9 \\
    \midrule
    \multirow{4}{*}{\small \sysname-1.5B}
     & 8 & 62.6 & 95.6 & 93.2 & 47.5 & 72.9 \\
     & 16 & 70.9 & 96.9 & 94.4 & 50.9 & 77.3 \\
     & 32 & 77.3 & 97.6 & 94.9 & 53.5 & 80.5 \\
     & 64 & 80.0 & 98.0 & 95.0 & 55.9 & 83.1 \\
    \midrule
    \multirow{4}{*}{\small Qwen2.5-Math-7B}
     & 8 & 39.5 & 86.1 & 80.1 & 32.5 & 50.9 \\
     & 16 & 46.2 & 90.0 & 84.7 & 39.3 & 58.2 \\
     & 32 & 51.6 & 92.8 & 89.1 & 45.4 & 64.1 \\
     & 64 & 56.7 & 94.8 & 95.0 & 50.4 & 68.7 \\
    \midrule
    \multirow{4}{*}{\small \sysname-7B}
     & 8 & 50.7 & 91.9 & 85.4 & 48.0 & 61.9 \\
     & 16 & 70.9 & 93.6 & 90.4 & 50.9 & 66.3 \\
     & 32 & 77.3 & 95.1 & 95.5 & 53.7 & 70.0 \\
     & 64 & 80.0 & 96.2 & 100.0 & 56.6 & 72.7 \\
    \bottomrule
  \end{tabular}
  \end{adjustbox}
  \caption{Additional results with pass@k metrics.}
\label{tab:passk}
\end{table}
Since our main experiments are reported using the pass@1 metric, we conducted additional pass@k experiments.
In these extra experiments, we sampled 64 samples to compute pass@k with $k=8/16/32/64$.
Results are reported in Table~\ref{tab:passk}.
As shown in the results table, both the base model and the trained model exhibit improved performance as the value of $k$ increases.
Consistent with the findings presented by~\citet{yue2025does}, reinforcement learning with verifiable rewards (RLVR) for LLMs primarily enhances the models' pass@1 performance.
In some cases, the pass@k performance of the RLVR-trained model even falls short of the base model.
For instance, DeepSeek-R1-Distill-Qwen-1.5B, trained using a large scale of high-quality CoT data distilled from DeepSeek-R1, shows limited improvement after training with RLVR in pass@1 performance in Table~\ref{tab:1}.
Even as $k$ increases, \sysname-1.5B's pass@k performance remains slightly inferior to the base model except for OlympiadBench.
However, \sysname-7B consistently outperforms its base model Qwen2.5-Math-7B for $k=1/8/16/32/64$ on all benchmarks, since Qwen2.5-Math-7B retains considerable potential for improvement.
\subsection{Training Analysis}
\label{appendix: traininganalysis}
\textbf{Training process analysis} \quad As illustrated by the training curves in Figure \ref{fig:2}, \sysname achieves optimization performance comparable to that of GPG, while exhibiting further improvements during the later steps of RL training.
In the Format Reward and Reward figures, \sysname consistently attains higher reward than GPG in the later steps of the training.
In addition, three Reward figures demonstrate significantly reduced fluctuations during the training process.
\sysname also outperforms GRPO.
Moreover, in the Response Length figure, our proposed \sysname demonstrates superior optimization, indicating better training effectiveness.
The final results presented in Table \ref{tab:1} demonstrate that our proposed \sysname achieves superior performance across five mathematical reasoning benchmarks.
To straightly and effectively analyze the training stability of \sysname, we calculated the variance of the training loss for \sysname as well as methods GRPO and GPG.
$\operatorname{Var}(\mathcal{L}_\mathrm{\sysname})=3.6\times 10^{-4}$, $\operatorname{Var}(\mathcal{L}_\mathrm{GPG})=3.9\times 10^{-4}$ and $\operatorname{Var}(\mathcal{L}_\mathrm{GRPO})=3.53\times 10^{-3}$.
The results show that \sysname exhibits the lowest variance, indicating relatively stable training.
Additionally, the results of the ablation experiments on $\operatorname{clip}$ operation demonstrate that \sysname can steadily improve model performance as training epochs progress.
\begin{figure}[b]
    \centering
    \begin{subfigure}[b]{0.23\textwidth}
        \centering
        \includegraphics[width=\textwidth]{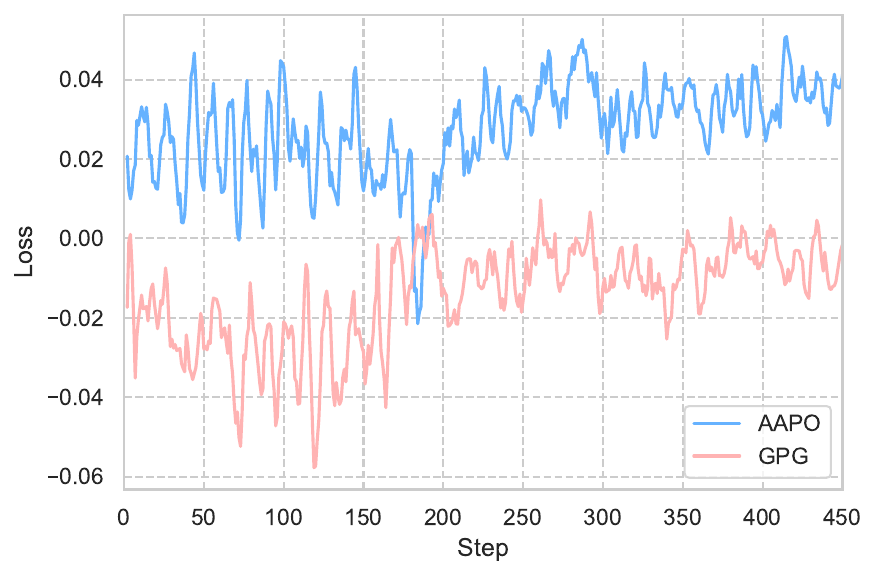}
        \label{fig:subfig9}
    \end{subfigure}%
    \hspace{0\textwidth}
    \begin{subfigure}[b]{0.23\textwidth}
        \centering
        \includegraphics[width=\textwidth]{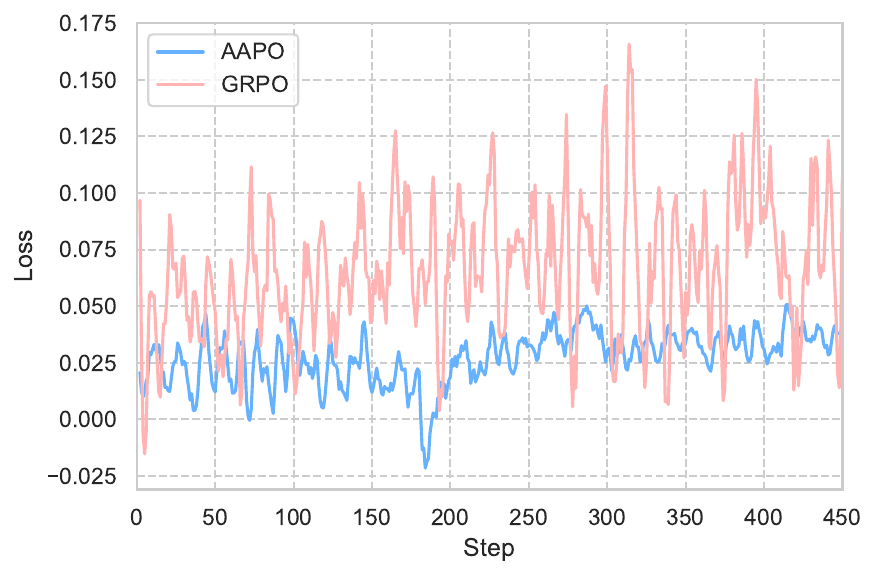}
        \label{fig:subfig10}
    \end{subfigure}%
    \caption{A comparative analysis of training loss between \sysname and GPG, \sysname and GRPO.}
    \label{fig:loss}
\end{figure}\\
\textbf{Training loss analysis} \quad We plot the loss curves of \sysname and GPG \citep{chu2025gpg} during training DeepSeek-R1-Distill-Qwen-1.5B on open-rs \citep{dang2025reinforcement} in Figure \ref{fig:loss}.
As observed, the loss values for \sysname remain predominantly positive throughout the training process. 
This indicates that \sysname primarily optimizes the policy by encouraging diverse responses better than the reference, assigning different gradient magnitudes according to the advantage estimated by $\hat{A}_{i,t}^*$, thus leading to performance improvements. 
In contrast, GPG exhibits mostly negative loss values, suggesting that it focuses on suppressing suboptimal responses as its main optimization strategy. 
It can be clearly seen from the Figure \ref{fig:loss} that the training process of \sysname is more stable than GRPO \citep{shao2024deepseekmath}.
These results of the analysis imply that models trained with our proposed \sysname demonstrate stronger generalization, resulting in superior performance as shown in Table \ref{tab:1}.
\section{Extra comparison with original results}
\label{sec: extra}
It is worth noting that evaluation results may be influenced by the computational device type.
We have also provided the original results of each model from the corresponding papers.
Here we present the original results reported in the corresponding papers for Qwen series models in Table~\ref{tab:5}.
\begin{table}[ht]
  \centering
  \small
  \setlength{\tabcolsep}{2pt}
  \begin{adjustbox}{width=0.48\textwidth, center} 
  \begin{tabular}{lccccc>{\columncolor{mygray}}c}
    \toprule
    Model&AIME24&MATH-500&AMC23&Minerva&OlympiadBench&Avg. \\
    \midrule
    \multicolumn{7}{c}{\textit{Qwen 1.5B Models}} \\
    \midrule
    {\small DeepSeek-R1-Distill-Qwen-1.5B} & 28.9 & 83.9 & -- & -- & -- & -- \\
    {\small GRPO-1.5B} \citep{dang2025reinforcement} & 46.7 & 84.4 & 72.5 & 26.8 & 51.3 & \underline{56.3} \\
    {\small GPG-1.5B \citep{chu2025gpg}} & 33.3 & 85.0 & 80.0 & 26.8 & 52.4 & 55.5 \\
    {\small Still-3-1.5B-Preview} \citep{chen2025empirical} & 39.3 & 85.5 & -- & -- & -- & -- \\
    \rowcolor{gray!15}
    {\small \sysname-1.5B (Ours)} & 33.3 & 86.0 & 80.0 & 30.9 & 53.3 & \textbf{56.7} \\
    \midrule
    \multicolumn{7}{c}{\textit{Qwen 7B Models}} \\
    \midrule
    {\small Qwen2.5-Math-7B \citep{yang2024qwen2math}} & -- & 55.4 & -- & -- & -- & -- \\
    {\small SimpleRL-Zero-7B \citep{Qwne2.5-7b-simplerl}} & 20.0 & 78.2 & 62.5 & 38.6 & 40.4 & 47.9 \\
    {\small GPG-7B\citep{chu2025gpg}} & 33.3 & 80.0 & 65.0 & 34.2 & 42.4 & 51.0 \\
    {\small OpenReasoner-Zero-7B \citep{hu2025open}} & -- & -- & -- & -- & -- & -- \\
    {\small Eurus-2-7B-PRIME \citep{cui2025process}} & 20.0 & 78.2 & 50.6 & 39.3 & 40.3 & 45.7 \\
    {\small Oat-Zero-7B\citep{liu2025understanding}} & 43.3 & 80.0 & 62.7 & 30.1 & 41.0 & \underline{51.4} \\
    \rowcolor{gray!15}
    {\small \sysname-7B (Ours)} & 30.0 & 82.4 & 70.0 & 35.3 & 44.3 & \textbf{52.4} \\
    \bottomrule
  \end{tabular}
  \end{adjustbox}
  \caption{Zero-shot pass@1 performance on mathematical reasoning benchmarks. All reported results in this table are directly adopted from the corresponding papers. Dashes (--) denote unavailable official score.}
\label{tab:5}
\end{table}
\section{Disclosure of AI assistants usage}
AI assistants are utilized to assist in the drafting, refinement, and wording adjustments of portions of the paper's text.
All content was ultimately reviewed and revised by the authors, who are solely responsible for the facts, assertions, and arguments presented herein.
\begin{table*}[ht]
  \centering
  \setlength{\tabcolsep}{2pt}
  \begin{adjustbox}{width=1\textwidth, center} 
  \begin{tabular}{lcccccc}
    \toprule
    MMLU task-domain&DeepSeek-R1-Distill-Qwen-1.5B&\sysname-1.5B&$\Delta$&Qwen2.5-Math-7B&\sysname-7B&$\Delta$ \\
    \midrule
    abstract\_algebra & 23.00 & 22.00 & -1.00 & 21.00 & 22.00 & +1.00 \\
    anatomy & 22.96 & 24.44 & +1.48 & 25.19 & 39.26 & +14.07 \\
    astronomy & 21.71 & 22.37 & +0.66 & 19.47 & 34.87 & +15.13 \\
    business\_ethics & 21.00 & 23.00 & +2.00 & 33.00 & 50.00 & +17.00 \\
    clinical\_knowledge & 30.19 & 28.68 & -1.51 & 23.02 & 40.00 & +16.98 \\
    college\_biology & 25.00 & 23.61 & -1.39 & 27.78 & 50.69 & +22.91 \\
    college\_chemistry & 29.00 & 26.00 & -3.00 & 21.00 & 27.00 & +6.00 \\
    college\_computer\_science & 25.00 & 25.00 & 0.00 & 27.00 & 37.00 & +10.00 \\
    college\_mathematics & 20.00 & 20.00 & 0.00 & 21.00 & 25.00 & +4.00 \\
    college\_medicine & 32.95 & 32.95 & 0.00 & 20.81 & 42.2 & +21.39 \\
    college\_physics & 22.55 & 21.57 & -0.98 & 21.57 & 27.45 & +5.88 \\
    computer\_security & 26.00 & 27.00 & +1.00 & 29.00 & 50.00 & +21.00 \\
    conceptual\_physics & 26.81 & 26.38 & -0.43 & 27.23 & 53.19 & +25.96 \\
    econometrics & 24.56 & 25.44 & +0.88 & 24.56 & 28.95 & +4.39 \\
    electrical\_engineering & 27.59 & 27.59 & 0.00 & 24.14 & 48.28 & +24.14 \\
    elementary\_mathematics & 21.16 & 20.90 & -0.26 & 21.16 & 31.48 & +10.32 \\
    formal\_logic & 24.60 & 25.40 & +0.80 & 28.57 & 35.71 & +7.14 \\
    global\_facts & 20.00 & 21.00 & +1.00 & 19.00 & 21.00 & +2.00 \\
    high\_school\_biology & 23.87 & 23.55 & -0.32 & 21.29 & 47.42 & +26.13 \\
    high\_school\_chemistry & 15.27 & 15.27 & 0.00 & 16.75 & 41.38 & +24.63 \\
    high\_school\_computer\_science & 26.00 & 26.00 & 0.00 & 28.00 & 51.00 & +23.00 \\
    high\_school\_european\_history & 29.70 & 30.30 & +0.60 & 21.81 & 40.00 & +18.19 \\
    high\_school\_geography & 30.30 & 28.79 & -1.51 & 22.73 & 50.51 & +27.78 \\
    high\_school\_government\_and\_politics & 27.46 & 26.94 & -0.52 & 30.57 & 46.63 & +16.06 \\
    high\_school\_macroeconomics & 27.95 & 27.95 & 0.00 & 22.56 & 42.82 & +20.26 \\
    high\_school\_mathematics & 21.85 & 21.85 & 0.00 & 21.11 & 22.59 & +1.48 \\
    high\_school\_microeconomics & 28.57 & 26.89 & -1.68 & 27.73 & 48.32 & +20.59 \\
    high\_school\_physics & 19.21 & 19.87 & +0.66 & 19.87 & 36.42 & +16.55 \\
    high\_school\_psychology & 24.40 & 24.40 & 0.00 & 25.69 & 57.43 & +31.74 \\
    high\_school\_statistics & 16.20 & 15.74 & -0.46 & 16.20 & 27.78 & +11.58 \\
    high\_school\_us\_history & 25.98 & 26.47 & +0.49 & 25.00 & 38.73 & +13.73 \\
    high\_school\_world\_history & 24.05 & 23.63 & -0.42 & 27.34 & 38.82 & +11.39 \\
    human\_aging & 18.83 & 20.18 & +1.35 & 33.63 & 38.57 & +4.94 \\
    human\_sexuality & 22.90 & 27.48 & +4.58 & 29.77 & 38.93 & +9.16 \\
    international\_law & 19.84 & 20.66 & +0.82 & 28.10 & 40.50 & +12.40 \\
    jurisprudence & 24.07 & 25.93 & +1.86 & 37.04 & 42.59 & +5.55 \\
    logical\_fallacies & 28.22 & 29.45 & +1.23 & 26.38 & 47.24 & +20.86 \\
    machine\_learning & 34.82 & 35.71 & +0.89 & 31.25 & 36.61 & +5.36 \\
    management & 18.45 & 18.45 & 0.00 & 25.24 & 48.54 & +23.30 \\
    marketing & 26.50 & 27.35 & +0.85 & 31.62 & 66.23 & +34.61 \\
    medical\_genetics & 20.00 & 23.00 & +3.00 & 30.00 & 33.00 & +3.00 \\
    miscellaneous & 23.24 & 23.88 & +0.64 & 32.70 & 52.11 & +19.41 \\
    moral\_disputes & 29.19 & 29.19 & 0.00 & 39.31 & 40.46 & +1.15 \\
    moral\_scenarios & 24.81 & 24.81 & 0.00 & 23.80 & 23.91 & +0.11 \\
    nutrition & 26.80 & 26.80 & 0.00 & 30.39 & 42.16 & +11.77 \\
    philosophy & 20.26 & 18.01 & -2.25 & 18.97 & 31.19 & +12.22 \\
    prehistory & 24.69 & 23.77 & -0.92 & 25.93 & 38.58 & +12.65 \\
    professional\_accounting & 23.76 & 24.11 & +0.35 & 23.76 & 30.14 & +6.38 \\
    professional\_law & 24.58 & 25.55 & +0.97 & 24.58 & 29.01 & +4.43 \\
    professional\_medicine & 16.18 & 15.81 & -0.37 & 18.38 & 27.57 & +9.19 \\
    professional\_psychology & 26.63 & 25.33 & -1.30 & 26.14 & 41.50 & +15.36 \\
    public\_relations & 17.27 & 20.91 & +3.64 & 22.73 & 35.46  & +12.73 \\
    security\_studies & 35.29 & 34.69 & -1.23 & 20.00 & 35.10 & +15.10 \\
    sociology & 25.37 & 28.86 & +3.49 & 27.86 & 44.78 & +16.92 \\
    us\_foreign\_policy & 26.00 & 25.00 & -1.00 & 39.00 & 50.00 & +11.00 \\
    virology & 19.88 & 21.08 & +1.20 & 28.92 & 42.17 & +13.25 \\
    world\_religions & 20.47 & 21.64 & +1.17 & 42.69 & 49.71 & +7.02 \\
    \midrule
    Average & 24.27 & 24.54 & +0.27 & 25.96 & 39.65 & +13.69\\
    \bottomrule
  \end{tabular}
  \end{adjustbox}
  \caption{Performance on MMLU benchmark.}
\label{tab:general}
\end{table*}
\end{document}